\documentclass[sigconf,screen]{acmart}

\usepackage{multibib}
\newcites{supp}{References}
\usepackage{multirow}
\usepackage{rotating}
\usepackage{booktabs}
\usepackage{colortbl}
\usepackage[ruled,lined,linesnumbered]{algorithm2e}
\SetKw{KwTo}{to}
\SetAlgoLongEnd 
\definecolor{mygray}{gray}{0.93}
\acmISBN{978-1-4503-XXXX-X/2018/06}

\setcopyright{none}
\renewcommand\footnotetextcopyrightpermission[1]{}
\begin{document}

\title{CouCE: A Unified Causal Framework for Debiased Deep Metric Learning}

\author{Xin Yuan}
\email{xinyuan@wust.edu.cn}
\affiliation{%
  \institution{Wuhan University of Science and Technology}
  \city{Wuhan}
  \country{China}}

\author{Zhenyang Niu}
\email{niuzy@wust.edu.cn}
\affiliation{%
  \institution{Wuhan University of Science and Technology}
  \city{Wuhan}
  \country{China}}

\author{Meiqi Wan}
\email{wanmeiqi@wust.edu.cn}
\affiliation{%
  \institution{Wuhan University of Science and Technology}
  \city{Wuhan}
  \country{China}}

\author{Huilin Zhu}
\email{zhuhuilin@wust.edu.cn}
\affiliation{%
  \institution{Wuhan University of Science and Technology}
  \city{Wuhan}
  \country{China}}

\author{Xin Xu}
\authornote{Corresponding author.}
\email{xuxin@wust.edu.cn}
\affiliation{%
  \institution{Wuhan University of Science and Technology}
  \city{Wuhan}
  \country{China}}

\author{Kui Jiang}
\email{jiangkui@hit.edu.cn}
\affiliation{%
  \institution{Harbin Institute of Technology}
  \city{Harbin}
  \country{China}}








\renewcommand{\shortauthors}{Trovato et al.}

\begin{abstract}
Deep Metric Learning (DML) often struggles with zero-shot generalization because standard objectives inherently capture what \textit{co-occurs} rather than what \textit{causes similarity}. Consequently, DML models are vulnerable to shortcut learning driven by two structurally distinct confounders: background spurious correlations (which create backdoor paths via scene context) and foreground nuisance perturbations (which inject non-semantic variations like pose or illumination). Although existing methods have proposed targeted solutions for each pathway individually, none can simultaneously address both due to their fundamentally distinct causal roles. To bridge this gap, we propose the Counterfactual Causal Embedding (CouCE), a unified causal framework that explicitly models and neutralizes both confounders. 
Specifically, we introduce Orthogonal Dictionary-Based Backdoor Adjustment (ODBA), which isolates spurious background patterns into a variance-gated dictionary and stably disentangles them from the learned embeddings via soft orthogonal regularization. Simultaneously, we propose Multi-Scale Randomized Causal Intervention (MSRCI) to enforce causal invariance against foreground nuisances through multi-scale Fourier amplitude randomization and a symmetric KL invariance constraint. Notably, CouCE seamlessly integrates with any proxy-based loss, incurring modest training overhead without requiring architectural modifications during inference. Extensive experiments on CUB-200-2011, Cars-196, and Stanford Online Products demonstrate that CouCE consistently achieves state-of-the-art performance, providing a principled and robust solution for debiased DML.

\end{abstract}





\maketitle

\begin{figure*}[t]
	\centering
	\includegraphics[width=\linewidth]{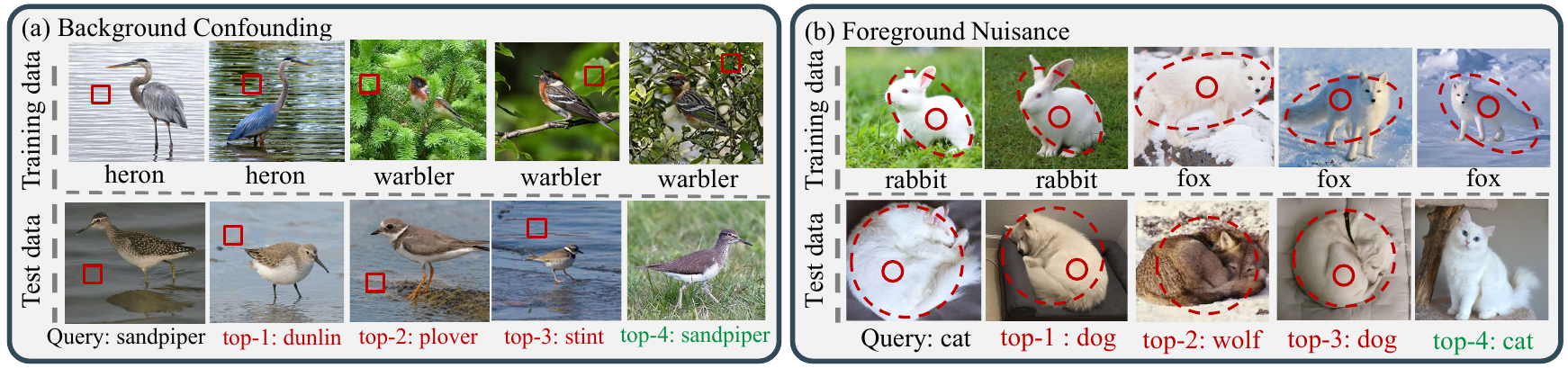}
    \caption{Two confounders in DML. \textbf{Red boxes} mark regions where background context misleads retrieval; \textbf{solid red circles} indicate foreground texture or color cues; \textbf{dashed red circles} indicate pose cues. (a)~\textbf{Background spurious correlation}: habitat co-occurrence causes a sandpiper query to retrieve dunlin, plover, and stint before the correct match. (b)~\textbf{Foreground nuisance perturbation}: pose and appearance biases cause a curled-up cat query to retrieve dogs and wolf before the correct cat.
    }
	\label{comparison}
    \vspace{-4mm}
\end{figure*}

\section{Introduction}\label{sec:intro}

Deep metric learning (DML) aims to learn an embedding space where distances reflect semantic similarity, serving as a cornerstone for image retrieval~\cite{yuan2023osap}, face recognition~\cite{xu2022rankinrank}, and fine-grained visual categorization~\cite{yuan2023searching}.
The DML research has advanced along two complementary fronts: pair-based losses~\cite{Hadsell2006DimensionalityRB,Schroff2015FaceNet,Wang2019MultiSimilarityLW} refine fine-grained intra-class structure but incur high sampling complexity, while proxy-based objectives~\cite{MovshovitzAttias2017NoFD,Kim2020ProxyAL,Qian2019SoftTripleLD,Peng2025ProxyAN,Silva2025PDLoss} introduce learnable class proxies to dramatically reduce training cost while preserving retrieval accuracy. 
In practice, however, the embeddings learned by these objectives can degenerate into low-dimensional subspaces and lose discriminative capacity, a phenomenon known as dimensional collapse~\cite{Roth2022NonisotropyRF,Jiang2024AntiCollapseLF}. 
To address this geometric vulnerability,  geometry-aware regularization methods~\cite{Roth2022NonisotropyRF,Jiang2024AntiCollapseLF} have emerged as a natural complement, enforcing non-isotropic, well-spread embeddings to prevent such collapse.
Together, these three families of methods have established a technically mature paradigm for DML, each contributing to improved encoding of observed co-occurrence patterns between samples and labels in the embedding space.
Despite significant advances in the above efforts, a fundamental yet underexplored vulnerability remains: standard objectives are inherently correlative, capturing what \textbf{\textit{co-occurs}} within the training distribution rather than what \textbf{\textit{causes similarity}}, and thus struggle with zero-shot generalization under distribution shifts~\cite{Musgrave2020RealityCheck,Patel2024ThreeThings,Arjovsky2019IRM}.

This vulnerability manifests as shortcut learning~\cite{Geirhos2020Shortcut,wang2022counterc} driven by two structurally distinct confounders, each introducing non-causal interference through a different mechanism. 
\textbf{The first is background spurious correlation:} dataset collection biases couple class labels with specific scenes (\textit{e.g.}, birds with wetlands), causing models to encode scene context as a discriminative cue and erroneously retrieve visually dissimilar objects sharing a similar background (Fig.~\ref{comparison}(a)). 
\textbf{The second is foreground nuisance perturbation:} object-level variations in pose, viewpoint, and illumination alter visual appearance without changing semantic identity, causing same-class samples to scatter in the embedding space along directions unrelated to semantics, and biasing retrieval toward appearance similarity rather than semantic identity (Fig.~\ref{comparison}(b)). 
Critically, these two confounders operate through fundamentally different causal roles, yet together they pose a dual threat to DML: background spurious correlations compromise retrieval precision, while foreground nuisances undermine recall.

To mitigate these confounders, existing research has advanced along two separate technical trajectories. \textbf{For background spurious correlation}, several approaches have been proposed. CaaM~\cite{wang2021causal} constructs a structural causal model for visual recognition and leverages causal intervention to learn background-invariant features, though it targets classification rather than metric learning. DCML~\cite{deng2022deep} introduces causal reasoning into DML, explicitly modeling background as a confounder and applying backdoor adjustment to remove spurious distances. BGAugment~\cite{kobs2022background} systematically analyzes background bias and proposes replacing backgrounds with random images as a simple augmentation strategy, yet this remains an observational data-level compensation rather than a causal intervention. 
\textbf{For foreground nuisance invariance}, advances come from domain generalization and self-supervised learning. FACT~\cite{Xu_2021_CVPR} employs Fourier amplitude randomization to simulate style variations, but applies it at a single resolution in pixel space and targets classification rather than metric ranking~\cite{wang2022sdadml}. ReLIC~\cite{Mitrovic2021ReLIC} formalizes data augmentations as interventions on style variables and enforces KL-divergence invariance, yet it is designed for self-supervised pretraining and does not address the inter-class ranking stability central to metric learning. 
Although these approaches demonstrate effectiveness in their respective domains, they remain confined to two disjoint technical paths: methods addressing the background pathway cannot simultaneously suppress foreground perturbations, while methods targeting style invariance are not designed for the causal intervention framework required by DML. Consequently, the dual deconfounding problem in DML remains unresolved.

To bridge this gap, we propose Counterfactual Causal Embedding (CouCE), a unified causal framework that explicitly models and neutralizes both confounders within a single architecture. CouCE is built around two complementary modules that introduce no architectural overhead during inference. Firstly, Orthogonal Dictionary-Based Backdoor Adjustment (ODBA), targets the background backdoor path~\cite{pearl2009causality} by maintaining a variance-gated EMA dictionary of background basis vectors and applying soft orthogonal regularization to drive sample embeddings out of the spurious background subspace, thereby separating background patterns from semantically discriminative features without spatial supervision. Secondly, Multi-Scale Randomized Causal Intervention (MSRCI), targets the foreground direct path by applying multi-scale Fourier amplitude randomization across feature maps to simulate appearance intervention, paired with a symmetric KL invariance constraint that ensures relative similarity rankings remain stable under foreground perturbation. Both modules are plug-and-play compatible with any proxy-based loss, incur modest training overhead, and introduce zero architectural overhead during inference. CouCE thus provides the first complete causal treatment of both structural interference pathways in DML, with zero inference overhead:

\begin{itemize}
    \item \textbf{Unified SCM and interventional objective.} We construct a structural causal model that formally distinguishes the background backdoor confounder from the foreground direct nuisance, and derive the interventional objective as a rigorous theoretical foundation for debiased DML.
    \item \textbf{CouCE with ODBA and MSRCI.} We propose two modules that incur no inference overhead: ODBA achieves backdoor adjustment via a variance-gated dictionary and soft orthogonal regularization; and MSRCI enforces causal invariance via multi-scale Fourier amplitude randomization and symmetric KL constraints.
    \item \textbf{State-of-the-art performance and generalizability.} 
    Extensive experiments on CUB-200-2011, Cars-196, and Stanford Online Products consistently demonstrate consistent superior outperformance of previous methods, with ablations further confirming the complementary nature of both modules across multiple proxy-based losses.
\end{itemize}

\section{Related Work}\label{sec:related}

\subsection{Deep Metric Learning}
Deep metric learning (DML) aims to learn an embedding space where semantically similar samples are pulled together while dissimilar ones are pushed apart, which can categorized into three groups. 
\textbf{(1) Pair-based losses:} Early DML methods rely on pairwise or tuplewise constraints. The contrastive loss~\cite{Hadsell2006DimensionalityRB,Kaya2019DeepML} directly penalizes distances between positive pairs and encourages separation of negative pairs. The triplet loss~\cite{Schroff2015FaceNet} extends this by comparing an anchor with a positive and a negative sample, requiring the distance to positive to be smaller than that to negative by a margin. Subsequent works introduce more sophisticated sampling strategies and loss formulations, such as lifted structured loss~\cite{Song2016SOP}, multi-similarity loss~\cite{Wang2019MultiSimilarityLW}, and margin-based variants~\cite{Wu2017SamplingMatters}, to better capture fine-grained intra-class and inter-class structures. However, these methods suffer from combinatorial sampling complexity, high sensitivity to hard-negative mining, and slow convergence on large-scale datasets. 
\textbf{(2) Proxy-based losses:} To address the sampling inefficiency, proxy-based objectives~\cite{MovshovitzAttias2017NoFD,Kim2020ProxyAL,Qian2019SoftTripleLD,Silva2025PDLoss,Maciag2026ProxyRobust} introduce a set of learnable class proxies as surrogate representatives. Instead of comparing a sample against all other samples, these methods compare it against a small number of proxies, dramatically reducing training complexity while preserving competitive retrieval accuracy. Proxy-NCA~\cite{MovshovitzAttias2017NoFD} uses proxies to approximate neighborhood component analysis, Proxy-Anchor~\cite{Kim2020ProxyAL} adopts an anchor-like formulation with adaptive margin scaling, and Proxy-AN~\cite{Peng2025ProxyAN} further refines the proxy assignment mechanism. SoftTriple~\cite{Qian2019SoftTripleLD} softens the hard proxy assignment to improve robustness. Despite their efficiency, proxy-based losses inherit the same geometric vulnerability as pair-based methods: the embeddings can collapse into low-dimensional subspaces. 
\textbf{(3) Geometry-aware regularization:} A distinct line of research identifies that embeddings learned by the above objectives can collapse into low-dimensional subspaces, losing discriminative capacity. This phenomenon is known as dimensional collapse. To mitigate this, geometry-aware regularization methods~\cite{Roth2022NonisotropyRF,Jiang2024AntiCollapseLF} explicitly encourage embeddings to be non-isotropic and well-spread across the embedding space, often by enforcing diversity constraints on the covariance matrix or promoting uniform angular distribution. These regularizers serve as a complementary component to base DML losses, addressing the collapse issue that pair-based and proxy-based methods commonly face. 
Despite these advances, all existing DML objectives remain fundamentally observational: they optimize similarity based on training co-occurrence statistics rather than causal structure. Consequently, they absorb spurious correlations and suffer from shortcut learning under distribution shifts, ultimately degrading zero-shot retrieval performance~\cite{Musgrave2020RealityCheck,Geirhos2020Shortcut,Patel2024ThreeThings,du2022invrl}.

\subsection{Non-Causal Interference Removal}
Existing efforts to mitigate non-causal interference have evolved along two separate trajectories corresponding to the two confounders identified in our causal analysis, \textit{i.e.}, \textbf{(1) background spurious correlation} and \textbf{(2) foreground nuisance invariance}. 
\textbf{For the former}, to address background confounding, several approaches have been proposed. Wang \textit{et al.}~\cite{wang2021causal} construct a structural causal model for visual recognition and leverage causal intervention to learn background-invariant features, though their work targets image classification rather than metric learning~\cite{qiu2021causalrec}. Deng \textit{et al.}~\cite{deng2022deep} introduce causal reasoning into DML for the first time, explicitly modeling background as a confounder and applying backdoor adjustment to remove spurious distances from the metric. Kobs \textit{et al.}~\cite{kobs2022background} systematically analyze background bias across DML benchmarks and propose a BGAugment strategy that replaces original backgrounds with random images, yet this remains an observational data-level compensation rather than a causal intervention~\cite{Bransby2024BackMix}. While these methods demonstrate progress, they address only the background pathway and do not extend to foreground nuisances. 
\textbf{For the latter}, foreground perturbations (\textit{e. g.}, pose, viewpoint, and illumination changes) relevant advances come from domain generalization and self-supervised learning. A key insight from signal processing is that Fourier amplitude primarily encodes style and illumination while phase preserves structural content~\cite{oppenheim1981importance,Piotrowski1982ADO,He2024FreqShortcutDG}. Building on this, Xu \textit{et al.}~\cite{Xu_2021_CVPR} propose a Fourier-based domain generalization framework using amplitude mix augmentation to simulate style variations, but their method operates at a single resolution in pixel space and targets classification rather than metric ranking. Mitrovic \textit{et al.}~\cite{Mitrovic2021ReLIC} formalize data augmentations as interventions on style variables and enforce KL-divergence invariance of predictions, yet their framework is designed for self-supervised pretraining and does not address the inter-class ranking stability central to DML~\cite{sun2021cdn}.
In summary, existing methods remain confined to two disjoint technical paths: those targeting background confounding cannot handle foreground perturbations, and vice versa. No prior work provides a unified causal framework for debised DML that simultaneously addresses both interference pathways within a single DML architecture, motivating our proposed CouCE.

\section{Method}\label{sec:method}

\begin{figure*}[t]
    \centering
    \includegraphics[width=\textwidth]{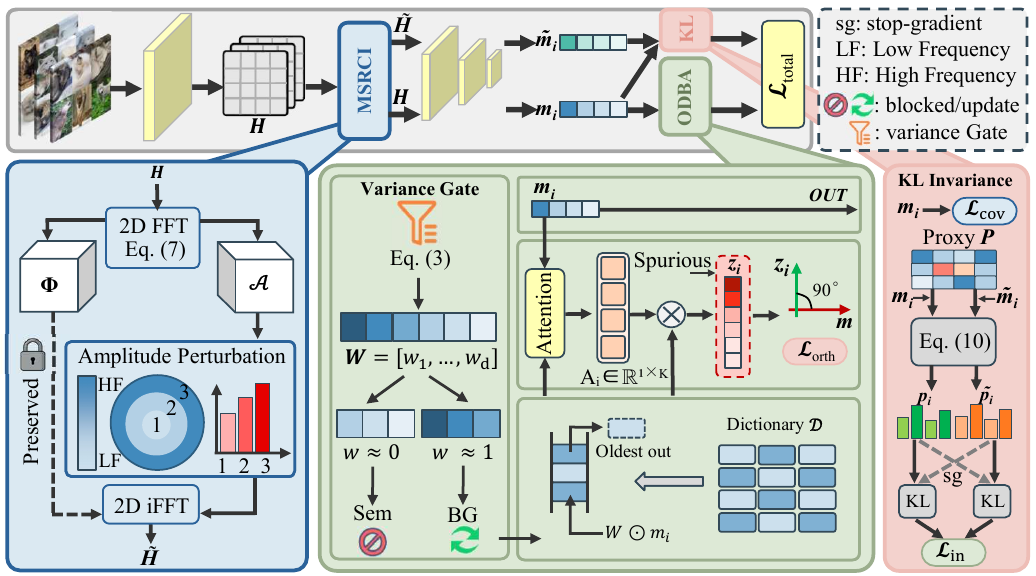}
    \caption{%
    \textbf{Overview of the CouCE framework.}
    Images pass through Stage\,1 to yield feature maps $H$.
    \textbf{MSRCI}: 2D FFT decomposes $H$ into preserved phase $\Phi$ and amplitude $\mathcal{A}$; $\mathcal{A}$ is perturbed across bands with increasing strength (LF$\to$HF) and iFFT yields $\tilde{H}$.
    Both views share downstream layers to produce $m_i$ and $\tilde{m}_i$.
    \textbf{ODBA}: a variance gate computes $W{=}[w_1,\ldots,w_d]$; semantic channels ($w{\approx}0$) are blocked while background channels ($w{\approx}1$) are enqueued via $W{\odot}m_i$ into FIFO dictionary $\mathcal{D}$; attention $A_i\in\mathbb{R}^{1\times K}$ retrieves spurious component $z_i$, and $\mathcal{L}_{\text{orth}}$ enforces $m_i\perp z_i$.
    \textbf{KL Invariance}: $m_i$ and $\tilde{m}_i$ are mapped via proxy $\boldsymbol{P}$ to distributions $p_i$, $\tilde{p}_i$; cross-sg symmetric KL yields $\mathcal{L}_{\text{inv}}$; $\mathcal{L}_{\text{cov}}$ decorrelates $m_i$ dimensions.
    }
    \label{fig:method_overview}
\end{figure*}

To address background spurious correlations and foreground nuisance perturbations, we propose CouCE, a unified causal framework with an SCM (Sec.~\ref{sec:scm}): ODBA blocks the background backdoor path via a variance-gated dictionary and soft orthogonal regularization (Sec.~\ref{sec:odba}), while MSRCI neutralizes the foreground direct path via multi-scale Fourier randomization and symmetric KL invariance (Sec.~\ref{sec:msrci}). Both modules are plug-and-play with any proxy-based loss and introduce zero inference overhead (Sec.~\ref{sec:integration}). The overview of the CouCE framework is shown in Fig.~\ref{fig:method_overview}

\begin{figure}[t]
    \centering
    \includegraphics[width=0.9\linewidth]{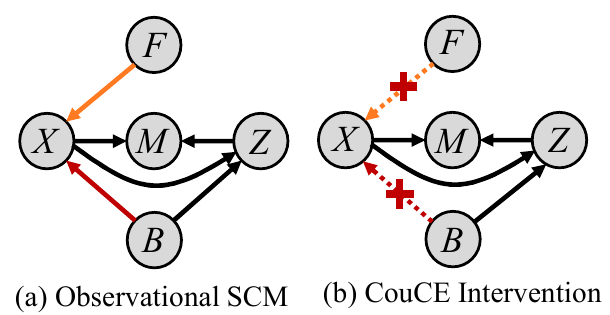}
    \caption{%
    \textbf{Structural Causal Model for DML.}
    \textbf{(a)}~Observational graph: The desired causal pathway is $X \!\to\! M$; two non-causal pathways contaminate it: the backdoor path $X \!\leftarrow\! B \!\to\! Z \!\to\! M$ (red) and the nuisance path $F \!\to\! X \!\to\! M$ (orange).
    \textbf{(b)}~Interventional graph after $do(X)$. $B \!\to\! X$ and $F \!\to\! X$ are severed (dashed with red crosses): ODBA realizes $do(X)$ by marginalizing over $B$, blocking the backdoor path $X \!\leftarrow\! B \!\to\! Z \!\to\! M$; MSRCI enforces invariance of $X \!\to\! M$ under $do(F)$.
    }
    \label{fig:causal_graph}
\end{figure}  

\subsection{Structural Causal Model for DML}\label{sec:scm}

\noindent \textbf{Problem Definition.}
Let $f_\theta: \mathcal{X} \to \mathbb{R}^d$ be the encoder minimizing $\mathcal{L}_{\text{DML}} = \mathbb{E}_{(x,y) \sim P_{\text{train}}}[\ell(f_\theta(x), y)]$. We construct the SCM~\cite{pearl2009causality} (Fig.~\ref{fig:causal_graph}) to identify two structurally distinct interference pathways that prior causal DML~\cite{deng2022deep} addresses only partially.

\noindent \textbf{Variables.}
The SCM contains five variables:
\textbf{$B$ (background environment):} dataset-level contextual prior; data collection biases create spurious correlations between $B$ and class identity (e.g., birds in wetlands).
\textbf{$F$ (foreground nuisance):} pose, viewpoint, illumination, and occlusion; semantically independent.
\textbf{$X$ (observed image):} jointly shaped by $B$ and $F$.
\textbf{$Z$ (spurious feature):} background co-occurrence patterns extracted by the encoder.
\textbf{$M$ (metric embedding):} final representation for distance computation.

\noindent \textbf{Causal Interventional Objective.}
As shown in Fig.~\ref{fig:causal_graph}(a), $B$ and $F$ jointly shape $X$; together they produce spurious features $Z$; and both $X$ and $Z$ determine $M$, conflating causal and non-causal signals.
Standard DML learns $P(M \mid X)$, contaminated by $B$ via the \emph{backdoor path} $X \leftarrow B \rightarrow Z \rightarrow M$.
Applying the backdoor criterion to $B$ yields the \emph{interventional distribution}:
\begin{equation}\label{eq:do_target}
    P\!\left(M \mid do(X)\right) = \sum_{b \in \mathcal{B}} P(M \mid X,\, B{=}b)\; P(b),
\end{equation}
This marginalizes $B$ under $P(b)$ rather than $P(b\!\mid\!X)$, severing the spurious background--class association. Yet the \emph{foreground nuisance path} $F\!\rightarrow\!X\!\rightarrow\!M$ persists: as a direct cause of $X$, $F$ requires a separate invariance constraint (Sec.~\ref{sec:msrci}).

\noindent \textbf{Computational obstacles and solution preview.}
Realizing Eq.~\eqref{eq:do_target} faces two obstacles:
\textbf{(I)}~the sum over $b$ is intractable because $B$ ranges over a continuous background space with no explicit enumeration or tractable density estimate;
\textbf{(II)}~the conditional $P(M \mid X, B{=}b)$ requires pairing each image with its background context, which is unavailable as supervision.
ODBA (Section~\ref{sec:odba}) resolves both obstacles jointly via a Normalized Weighted Geometric Mean approximation and a dynamic background dictionary; MSRCI (Section~\ref{sec:msrci}) addresses the foreground path through multi-scale Fourier amplitude randomization in feature space.

\subsection{Orthogonal Dictionary-Based Backdoor Adjustment}\label{sec:odba}

To address the two computational obstacles of realizing the backdoor adjustment in Eq.~\eqref{eq:do_target} without spatial annotations, we propose \textbf{Orthogonal Dictionary-Based Backdoor Adjustment (ODBA)}. ODBA maintains a dynamic dictionary of background-dominant feature patterns and imposes a soft orthogonality constraint between clean embeddings and the spurious subspace.

We first address obstacle~(I) via the \emph{Normalized Weighted Geometric Mean} (NWGM) approximation~\cite{yang2021deconfounded}:
\begin{equation}\label{eq:nwgm}
    P(M \mid do(X)) \approx \text{Softmax}\!\left[\mathbb{E}_{b}\!\left[g(z_x,\, z_b)\right]\right],
\end{equation}
where $g(z_x, z_b)$ is the proxy similarity score between the semantic feature $z_x$ (the clean embedding $m_i$) and background context $z_b$; $\ell_2$-normalization at enqueue makes $g$ strictly linear in $z_b$ (see Appendix~\ref{app:nwgm}), validating the approximation.
Approximating $P(b)$ with $K$ dictionary atoms yields $\mathbb{E}_b[g(z_x,z_b)] \approx \sum_k A_k \cdot g(z_x, \mathcal{D}_k)$ with $A_k \propto \exp(z_x \cdot \mathcal{D}_k^\top / \sqrt{d})$, collapsing the intractable marginalization into a differentiable attention operation while the dictionary implicitly represents $P(b)$, resolving both obstacles simultaneously.

Without filtering, a naive dictionary risks absorbing class-shared semantic commonalities alongside background patterns. We therefore introduce a \textbf{variance-gated EMA update}: in each mini-batch, we compute per-channel within-class variance $\text{Var}_{\text{intra}}$ and between-class variance $\text{Var}_{\text{inter}}$, and define the gating weight:
\begin{equation}\label{eq:var_gate}
    w_c = \sigma\!\left(\frac{\text{Var}_{\text{intra}}(c)}{\text{Var}_{\text{inter}}(c) + \epsilon} - \tau_g\right) \cdot \mathbb{1}\!\left[\text{Var}_{\text{intra}}(c)+\text{Var}_{\text{inter}}(c)>\delta\right],
\end{equation}
where $\sigma$ is the sigmoid function, $\tau_g$ is a threshold, and $\mathbb{1}[\cdot]$ screens out inert channels ($\delta = 10^{-4}$); since per-batch estimates are noisy, the variance ratio is smoothed via EMA (momentum $\mu$) before applying the gate, ensuring stable channel selection.
Background channels (small $\text{Var}_{\text{inter}}$, high $\text{Var}_{\text{intra}}$) receive $w_c \approx 1$ and are admitted; semantically discriminative channels receive $w_c \approx 0$ and are suppressed.
The dictionary is maintained as a FIFO queue of capacity $K$. Let $m_i \in \mathbb{R}^d$ denote the clean embedding produced by the embedding head $f_{\rm emb}$ after global average pooling. The gated and $\ell_2$-normalized features are enqueued at each step:
\begin{equation}\label{eq:enqueue}
    \mathcal{D} \leftarrow \textsc{Enqueue}\!\left(\mathcal{D},\; \left\{ \frac{W \odot m_i}{\|W \odot m_i\|_2} \right\}_{i=1}^{N}\right),
\end{equation}
where $W = [w_1, \ldots, w_d]$ is the vector of per-channel gating weights $w_c$ defined in Eq.~\eqref{eq:var_gate}, and oldest entries are dequeued at capacity.

For a clean embedding $m_i \in \mathbb{R}^d$, we compute a conditional background distribution via scaled dot-product attention:
\begin{equation}\label{eq:attn}
    A_i = \text{Softmax}\!\left(\frac{m_i \cdot \mathcal{D}^\top}{\sqrt{d}}\right) \in \mathbb{R}^{1 \times K},
\end{equation}
where $A_i$ approximates $P(b \mid m_i)$, the posterior probability of each background atom given the observation; the spurious feature is then reconstructed as $z_i = A_i \cdot \mathcal{D} \in \mathbb{R}^d$, realizing the finite-sample approximation of Eq.~\eqref{eq:nwgm}, without explicit supervision.

Hard orthogonal projection ($m_{\text{deconf}} = m_i - \text{proj}_{z_i} m_i$) suffers from gradient instability and may erroneously remove semantic components when the variance gate is imperfect. We instead impose orthogonality as a differentiable regularization loss:
\begin{equation}\label{eq:ortho_loss}
    \mathcal{L}_{\text{orth}} = \frac{1}{N} \sum_{i=1}^{N} \left| \frac{\langle m_i,\, z_i \rangle}{\|m_i\|_2 \cdot \|z_i\|_2 + \varepsilon} \right|^2.
\end{equation}
We detach $z_i$ from the computational graph so that gradients flow only through $m_i$, preventing the attention mechanism from trivially satisfying the constraint without the backbone actually suppressing background components; when $\mathcal{L}_{\text{orth}} \to 0$, the embedding becomes orthogonal to the spurious subspace $\mathcal{S}_B = \text{span}(\mathcal{D})$, realizing the backdoor adjustment that neutralizes $B$-mediated confounding.

\subsection{Multi-Scale Randomized Causal Intervention}\label{sec:msrci}

While ODBA blocks the background backdoor path, the foreground nuisance path $F \rightarrow X \rightarrow M$ remains active. Since $F$ is a direct cause of $X$ rather than a confounder, neutralizing it requires simulating $do(F)$ to enforce representational invariance under foreground changes. To this end we propose \textbf{Multi-Scale Randomized Causal Intervention (MSRCI)}, which intervenes directly in feature space via multi-scale Fourier amplitude randomization as a proxy for $do(F)$, paired with a symmetric KL invariance constraint on proxy similarity distributions to stabilize metric rankings.

Let $H \in \mathbb{R}^{N \times C \times W' \times H'}$ denote the intermediate feature map (e.g., after \texttt{layer3} in ResNet); since the phase spectrum encodes geometric structure while the amplitude spectrum encodes style and illumination, we simulate $do(F{=}\text{random})$ by randomizing amplitude across multiple frequency bands while strictly preserving phase.
We decompose each feature channel $H_c$ via 2D FFT:
\begin{equation}\label{eq:fft}
    \mathcal{F}(H_c) = \text{FFT2D}(H_c) = \mathcal{A}_c \cdot e^{j\Phi_c},
\end{equation}
where $\mathcal{A}_c$ encodes appearance and $\Phi_c$ preserves structure.

Uniform perturbation across all bands would either corrupt semantic content encoded in low frequencies or under-perturb nuisance-carrying high frequencies; we therefore partition the amplitude spectrum into $S$ concentric annular frequency bands with monotonically increasing perturbation strengths:
\begin{equation}\label{eq:amp_perturb}
    \tilde{\mathcal{A}}_c = \mathcal{A}_c \cdot
    \left(1 + \sum_{s=1}^{S} \mathbb{1}_{R_s}(\omega) \cdot \gamma_s \cdot \Delta A_s \right),
\end{equation}
where $R_s$ is the mask for band $s$, $\Delta A_s \sim \text{Uniform}(-1, 1)$, and $\gamma_s < 1$ increases monotonically so that low-frequency bands receive small perturbations to preserve global semantic structure such as object shape while high-frequency bands receive large perturbations to cover the full range of $F$-induced nuisances.
The intervened map is reconstructed via inverse FFT (channel-wise):
\begin{equation}\label{eq:ifft}
    \tilde{H}_c = \text{iFFT2D}\!\left(\tilde{\mathcal{A}}_c \cdot e^{j\Phi_c}\right),
\end{equation}
yielding the full intervened feature map $\tilde{H}$.
operationalizing $do(F)$ in our SCM: randomizing amplitude draws $F$ from its marginal while preserving phase holds semantic structure constant, realizing the graph-surgery effect without requiring an explicit partition of the input into semantic and nuisance components.

The maps $H$ and $\tilde{H}$ are processed by the backbone's remaining stages, GAP, and the shared embedding head $f_{\text{emb}}$, yielding clean embedding $m_i$ and intervened embedding $\tilde{m}_i$. Rather than naive L2 consistency, we require \emph{functional invariance}: consistent ranking across all proxies. We define proxy similarity distributions:
\begin{equation}\label{eq:proxy_dist}
    p_i = \text{Softmax}\!\left(\frac{m_i \cdot P^\top}{\tau}\right), \quad
    \tilde{p}_i = \text{Softmax}\!\left(\frac{\tilde{m}_i \cdot P^\top}{\tau}\right),
\end{equation}
and impose invariance via the symmetric KL divergence with stop-gradient to prevent distributional collapse:
\begin{equation}\label{eq:inv}
    \mathcal{L}_{\text{inv}} = \frac{1}{2N} \sum_{i=1}^{N}
    \left[ D_{\text{KL}}\!\left(p_i \,\|\, \text{sg}(\tilde{p}_i)\right)
    + D_{\text{KL}}\!\left(\tilde{p}_i \,\|\, \text{sg}(p_i)\right) \right],
\end{equation}
where $\text{sg}(\cdot)$ denotes the stop-gradient operator~\cite{Chen2021SimSiam} that prevents both branches from collapsing to a trivial uniform distribution.

To mirror the causal independence of $F$ from semantic identity in representation space and prevent amplitude perturbations from bleeding into semantic dimensions, we add a covariance penalty~\cite{Bardes2022VICReg}. Let $M \in \mathbb{R}^{N\times d}$ be the matrix whose rows are the clean embeddings of the current mini-batch; we regularize via
\begin{equation}\label{eq:cov}
    \mathcal{L}_{\text{cov}} = \frac{1}{d} \sum_{i \neq j}
    \left[\text{Cov}(M_i, M_j)\right]^2,
\end{equation}
encouraging dimensional independence across embeddings~\cite{Scholkopf2021Toward,Komanduri2024ICMVAE}.

\begin{table*}[t]
\caption{Performance comparison with with state-of-the-art DML methods on CUB-200-2011, Cars-196, and Stanford Online Products. 
The top part indicates that the backbone uses BN-Inception. The bottom part denotes that the backbone uses ResNet-50. 
We report R@1, R@2, RP, MAP@R, and NMI (\%). 
\textbf{Bold} indicates best performance.}
\label{tab:merged_comparison}
\centering
\resizebox{\textwidth}{!}{%
\begin{tabular}{llccrrrrrcrrrrrcrrrrr}
\toprule
\multirow{2}{*}{} & \multirow{2}{*}{\textbf{Method}} & \multirow{2}{*}{\textbf{Venue}} && \multicolumn{5}{c}{\textbf{CUB-200-2011}} && \multicolumn{5}{c}{\textbf{Cars-196}} && \multicolumn{5}{c}{\textbf{SOP}} \\
\cmidrule(lr){5-9} \cmidrule(lr){11-15} \cmidrule(lr){17-21}
&&&& \textbf{R@1} & \textbf{R@2} & \textbf{RP} & \textbf{MAP@R} & \textbf{NMI} && \textbf{R@1} & \textbf{R@2} & \textbf{RP} & \textbf{MAP@R} & \textbf{NMI} && \textbf{R@1} & \textbf{R@2} & \textbf{RP} & \textbf{MAP@R} & \textbf{NMI} \\
\midrule

\multirow{12}{*}{\rotatebox{90}{\small BN-Inception}}
& HIST~\cite{Lim2022HypergraphInducedST} & CVPR'22 && 67.48 & 79.02 & 33.13 & 23.54 & 70.69 && 86.31 & 91.94 & 34.77 & 25.52 & 72.61 && 78.03 & 89.40 & 51.31 & 48.36 & 90.45 \\
& DAS~\cite{Liu2022DASDS} & ECCV'22 && 64.89 & 77.70 & 32.18 & 22.95 & 70.24 && 84.85 & 90.80 & 33.56 & 24.01 & 70.44 && 77.02 & 89.41 & 50.79 & 47.67 & 90.03 \\
& NIR~\cite{Roth2022NonisotropyRF} & CVPR'22 && 67.76 & 79.16 & 34.30 & 23.85 & 71.02 && 86.84 & 92.22 & 35.35 & 25.45 & 73.52 && 78.30 & 89.27 & 51.48 & 48.34 & 90.24 \\
& PLG~\cite{Roth2022IntegratingLG} & CVPR'22 && 68.25 & 79.92 & 34.12 & 24.33 & 71.52 && 86.55 & 92.14 & 35.21 & 25.81 & 71.15 && 78.02 & 90.12 & 51.48 & 48.50 & 89.98 \\
& DCML~\cite{deng2022deep} & ICML'22 && 67.73 & 79.86 & 35.99 & 25.28 & 72.08 && 86.36 & 91.50 & 36.38 & 27.41 & 72.56 && 77.84 & 89.46 & 51.32 & 48.06 & 90.10 \\
& ACL~\cite{Jiang2024AntiCollapseLF} & TMM'24 && 68.83 & 80.11 & 35.73 & 24.95 & 71.29 && 88.22 & 92.59 & 37.77 & 28.74 & 73.08 && 76.43 & 88.59 & 50.49 & 47.42 & 89.50 \\
& DADA~\cite{Ren2024TowardsIP} & AAAI'24 && 67.89 & 79.84 & 35.66 & 25.72 & 72.01 && 88.25 & 91.64 & 37.69 & 28.95 & 73.16 && 79.31 & 90.98 & 51.91 & 48.70 & 90.40 \\
& MFT~\cite{furusawa2024mean} & ICLR'24 && 67.66 & 78.89 & 36.57 & 26.08 & 68.04 && 83.32 & 90.83 & 33.17 & 23.04 & 67.45 && 76.97 & 89.33 & 51.36 & 48.15 & 89.60 \\
& DDML~\cite{park2025deep} & AAAI'25 && 67.71 & 79.23 & 35.68 & 25.12 & 71.71 && 86.88 & 91.87 & 36.35 & 27.39 & 72.68 && 78.93 & 90.26 & 51.94 & 48.82 & 90.30 \\
& PFML~\cite{Bhatnagar2024PotentialFB} & CVPR'25 && 69.13 & 80.38 & 36.53 & 25.93 & 71.32 && 88.85 & 93.24 & 38.56 & 29.65 & 73.93 && 79.27 & 90.40 & 51.89 & 48.96 & 90.50 \\
\cmidrule(lr){2-21}
\rowcolor{mygray} & \textbf{CouCE} & \textbf{Ours} && \textbf{70.95} & \textbf{80.90} & \textbf{38.35} & \textbf{27.45} & \textbf{72.15} && \textbf{89.50} & \textbf{93.80} & \textbf{40.05} & \textbf{31.70} & \textbf{74.65} && \textbf{80.61} & \textbf{91.62} & \textbf{53.48} & \textbf{50.46} & \textbf{90.92} \\

\midrule

\multirow{14}{*}{\rotatebox{90}{\small ResNet-50}}
& HIST~\cite{Lim2022HypergraphInducedST} & CVPR'22 && 69.09 & 80.01 & 34.08 & 24.24 & 72.03 && 88.28 & 93.03 & 35.69 & 26.02 & 73.73 && 79.89 & 90.34 & 53.11 & 50.06 & 91.40 \\
& DAS~\cite{Liu2022DASDS} & ECCV'22 && 67.35 & 78.53 & 33.18 & 23.68 & 71.21 && 87.06 & 92.60 & 34.24 & 24.71 & 72.09 && 79.24 & 90.14 & 52.64 & 49.58 & 91.00 \\
& NIR~\cite{Roth2022NonisotropyRF} & CVPR'22 && 68.49 & 79.86 & 34.94 & 24.04 & 72.09 && 87.80 & 92.75 & 35.76 & 25.97 & 74.53 && 79.88 & 90.34 & 52.88 & 50.02 & 90.76 \\
& PLG~\cite{Roth2022IntegratingLG} & CVPR'22 && 69.22 & 80.14 & 34.49 & 24.86 & 73.09 && 88.68 & 93.96 & 37.43 & 27.37 & 72.21 && 80.38 & 91.01 & 53.27 & 50.07 & 90.14 \\
& IDML~\cite{Wang2023IntrospectiveDM} & TPAMI'23 && 68.56 & 79.45 & 35.88 & 25.21 & 73.35 && 89.79 & 93.66 & 39.35 & 30.32 & 76.94 && 80.45 & 91.11 & 53.22 & 50.01 & 91.49 \\
& HSE~\cite{Yang2023HSEHS} & ICCV'23 && 68.45 & 79.85 & 35.75 & 24.97 & 71.42 && 88.33 & 93.60 & 38.55 & 29.62 & 75.17 && 79.51 & 90.90 & 52.82 & 49.63 & 90.90 \\
& HIER~\cite{Kim2023HIERML} & CVPR'23 && 68.09 & 78.73 & 35.40 & 24.52 & 71.25 && 87.49 & 92.36 & 37.02 & 27.87 & 73.47 && 79.87 & 91.50 & 52.89 & 49.98 & 90.80 \\
& ACL~\cite{Jiang2024AntiCollapseLF} & TMM'24 && 69.73 & 80.83 & 36.26 & 25.66 & 73.39 && 89.55 & 94.13 & 39.51 & 30.16 & 75.63 && 80.20 & 91.51 & 53.29 & 50.09 & 91.10 \\
& DADA~\cite{Ren2024TowardsIP} & AAAI'24 && 70.69 & 81.68 & 37.23 & 27.31 & 72.84 && 91.21 & 94.54 & 40.48 & 31.58 & 76.90 && 80.36 & 91.67 & 53.35 & 50.48 & 91.30 \\
& MFT~\cite{furusawa2024mean} & ICLR'24 && 67.92 & 79.11 & 37.12 & 26.73 & 68.58 && 84.44 & 90.87 & 33.73 & 23.44 & 67.82 && 77.63 & 90.01 & 51.53 & 48.33 & 89.70 \\
& DDML~\cite{park2025deep} & AAAI'25 && 69.39 & 80.85 & 36.12 & 25.66 & 72.80 && 88.90 & 93.14 & 38.25 & 29.26 & 75.67 && 81.13 & 91.43 & 53.92 & 50.97 & 91.70 \\
& PFML~\cite{Bhatnagar2024PotentialFB} & CVPR'25 && 71.01 & 81.67 & 38.10 & 28.28 & 73.24 && 91.40 & 94.85 & 40.93 & 31.88 & 77.39 && 81.15 & 91.79 & 53.95 & 50.76 & 91.70 \\
\cmidrule(lr){2-21}
\rowcolor{mygray} & \textbf{CouCE} & \textbf{Ours} && \textbf{73.23} & \textbf{82.24} & \textbf{40.37} & \textbf{30.03} & \textbf{74.23} && \textbf{92.73} & \textbf{95.53} & \textbf{42.69} & \textbf{34.36} & \textbf{78.24} && \textbf{82.34} & \textbf{92.47} & \textbf{55.82} & \textbf{52.75} & \textbf{92.16} \\

\bottomrule
\end{tabular}%
}
\end{table*}

\subsection{Integration and Training}\label{sec:integration}

\paragraph{Data flow and module placement.}
As shown in Fig.~\ref{fig:method_overview}, MSRCI operates on the intermediate feature map $H$ (between Stage~1 and Stage~2) to produce $\tilde{H}$; both views pass through the shared backbone tail to yield embeddings $m$ and $\tilde{m}$, after which ODBA operates on $m$. Intervening at this stage preserves the spatial frequency structure required for Fourier decomposition while targeting a semantically meaningful representation for backdoor adjustment.

\paragraph{Training objective.}
The total loss combines the proxy-based DML objective with three causal regularizers:
\begin{equation}\label{eq:total_loss}
    \mathcal{L}_{\text{total}}
    = \mathcal{L}_{\text{DML}}(m)
    + \lambda_1 \cdot \mathcal{L}_{\text{orth}}
    + \lambda_2 \cdot \mathcal{L}_{\text{inv}}
    + \lambda_3 \cdot \mathcal{L}_{\text{cov}},
\end{equation}
where $\lambda_1$, $\lambda_2$, $\lambda_3$ weight spurious alignment, foreground invariance, and dimensional independence penalties, respectively (sensitivity analyses in Section~\ref{sec:ablation}). The three regularizers are non-conflicting by design: $\mathcal{L}_{\text{orth}}$ and $\mathcal{L}_{\text{inv}}$ target orthogonal interference pathways, while $\mathcal{L}_{\text{cov}}$ prevents their interaction from introducing dimensional redundancy into the learned embeddings. All three converge stably within a single training schedule. At test time, the MSRCI branch and ODBA dictionary are discarded, with no architectural modification, leaving zero inference overhead (Appendix~\ref{app:algorithm}).

\section{Experiments}\label{sec:exp}

\subsection{Experimental Setup}\label{sec:setup}

\textbf{Datasets.}
We evaluate on three standard DML benchmarks:
(i)~\textbf{CUB-200-2011}~\cite{WahCUB200} contains 11,788 images of 200 bird species, making it a widely used fine-grained dataset. The first 100 classes (5,864 images) are used for training.
(ii)~\textbf{Cars-196}~\cite{Krause2013Cars196} contains 16,185 images of 196 car models with significant variations in pose, lighting, and background. The first 98 classes (8,054 images) are used for training.
(iii)~\textbf{Stanford Online Products (SOP)}~\cite{Song2016SOP} contains 120,053 images in 22,634 product classes for large-scale retrieval scenario, 11,318 classes (59,551 images) are used for training.

\noindent \textbf{Backbones.}
Following standard protocol, we evaluate with BN-Inception~\cite{Ioffe2015BN} (pretrained on ImageNet) and ResNet-50~\cite{He2016ResNet} (pretrained on ImageNet). 
Both backbones are used without additional modifications.
The embedding dimension is set to 512.

\noindent \textbf{Training details.}
We use Proxy-AN~\cite{Peng2025ProxyAN} as the default base DML loss.
The backbone is optimized with Adam ($\text{lr}=10^{-4}$, weight decay $10^{-4}$); proxies use a separate Adam ($\text{lr}=10^{-2}$).
Images are resized to $256 \times 256$ and randomly cropped to $224 \times 224$ with a random horizontal flip.
ODBA: dictionary capacity $K{=}2048$, variance-ratio EMA momentum $\mu{=}0.999$, gating threshold $\tau_g{=}1.0$, inert-channel cutoff $\delta{=}10^{-4}$.
MSRCI: $S{=}3$ bands, $\gamma_s \in \{0.2, 0.4, 0.8\}$, $\tau{=}0.1$.
Loss coefficients: $\lambda_1{=}0.05$, $\lambda_2{=}0.1$, $\lambda_3{=}0.04$.
All experiments run for 80 epochs on a single RTX 2080Ti; results are averaged over 3 seeds.
The MSRCI intervention branch is active only during training.

\noindent \textbf{Evaluation metrics.}
Following Musgrave~\textit{et al.}~\cite{Musgrave2020RealityCheck}, we report Recall@$k$ ($k{=}1, 2$), R-Precision (RP), MAP@R, and NMI, emphasising MAP@R for full-ranking quality.

\begin{figure*}[t]
    \centering
    \includegraphics[width=\textwidth]{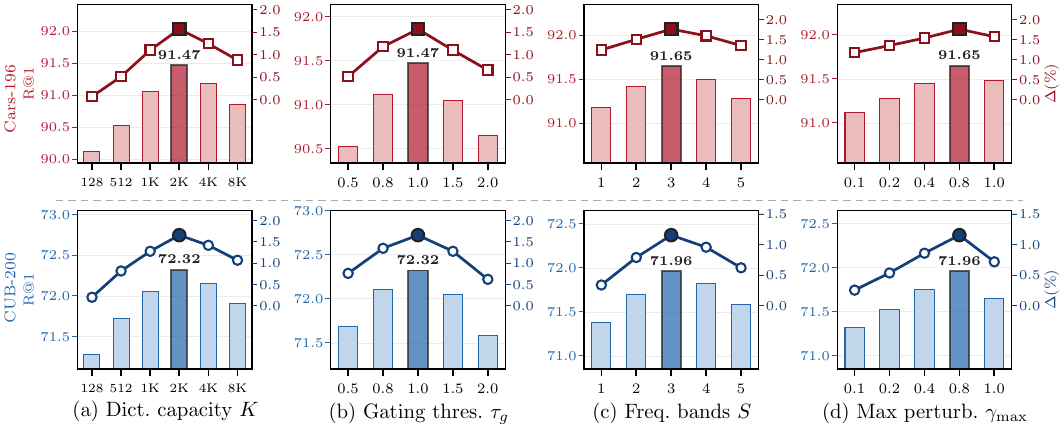}
    \caption{%
    Ablation results of core continuous hyperparameters on CUB-200-2011 and Cars-196 (ResNet-50).
    From left to right:
    \textbf{(a)}~dictionary capacity $K$,
    \textbf{(b)}~variance-gating threshold $\tau_g$,
    \textbf{(c)}~number of frequency bands $S$,
    \textbf{(d)}~max intervention strength $\gamma_{\max}$.
    }
    \label{fig:hyper_sensitivity}
\end{figure*}

\subsection{Comparison with State-of-the-Art}\label{sec:sota}

Table~\ref{tab:merged_comparison} presents a comprehensive comparison with recent DML methods on CUB-200-2011, Cars-196, and Stanford Online Products.

\textbf{Results.}
On \textbf{CUB-200-2011}, where background confounding is severe (habitat--species co-occurrence), CouCE achieves R@1 of 70.95\% (BN-Inception) and 73.23\% (ResNet-50), surpassing the previous best PFML by +1.82\% and +2.22\% respectively, with notably larger gains on RP (+2.27\% on ResNet-50), and consistent improvements on MAP@R across both backbones.
On \textbf{Cars-196}, where viewpoint variation and semantic absorption (shared tyres/chassis) are the main challenges, CouCE reaches 92.73\% R@1 (ResNet-50), outperforming PFML (91.40\%) by +1.33\% R@1 and +2.48\% MAP@R. The larger gain on MAP@R than on R@1 is consistent with the role of MSRCI: nuisance invariance stabilizes rankings beyond the top-1 position, improving full-ranking quality more broadly than precision at the single nearest neighbor.
On \textbf{SOP}, CouCE achieves 82.34\% R@1 (ResNet-50), surpassing PFML by +1.19\% R@1 and +1.99\% MAP@R. The improvement margin is more moderate than on CUB and Cars, consistent with the SCM analysis: SOP images are product photographs typically captured against clean backgrounds, reducing the severity of the background confounding that ODBA targets; the primary source of interference is intra-class viewpoint and lighting variation, which MSRCI addresses. All five metrics showed small but steady gains. This proves CouCE can handle large-scale retrieval and stays effective even if one interference pathway is weak.
Compared with the most directly comparable causal baseline DCML~\cite{deng2022deep}, CouCE shows consistent advantages across all datasets, attributable to addressing both confounders simultaneously and avoiding gradient instability from sample re-weighting (a detailed analysis is provided in the supplementary material).
The improvement pattern is consistent with our causal analysis: CUB benefits most from ODBA (background deconfounding), Cars from MSRCI (nuisance invariance), while SOP shows moderate but consistent gains.

\begin{table}[t]
\caption{Ablation study results of variance-ratio EMA momentum $\mu$ on CUB and Cars (ResNet-50, PA+ODBA).}
\centering
\small
\resizebox{\columnwidth}{!}{%
\setlength{\tabcolsep}{3pt}
\begin{tabular}{lccccccc}
\toprule
\multirow{2}{*}{\textbf{$\mu$}} & \multicolumn{3}{c}{\textbf{CUB}} && \multicolumn{3}{c}{\textbf{Cars}} \\
\cmidrule(lr){2-4} \cmidrule(lr){6-8}
& \textbf{R@1} & \textbf{M@R} & \textbf{NMI} && \textbf{R@1} & \textbf{M@R} & \textbf{NMI} \\
\midrule
0.9      & 71.55 & 28.72 & 73.63 && 90.52 & 30.48 & 77.22 \\
0.99     & 71.92 & 29.02 & 73.72 && 90.95 & 30.85 & 77.42 \\
0.995    & 72.15 & 29.22 & 73.78 && 91.22 & 31.08 & 77.55 \\
\rowcolor{mygray}
\textbf{0.999 (default)} & \textbf{72.32} & \textbf{29.37} & \textbf{73.81} && \textbf{91.47} & \textbf{31.26} & \textbf{77.65} \\
0.9999   & 72.08 & 29.10 & 73.74 && 91.08 & 30.90 & 77.45 \\
\bottomrule
\end{tabular}%
}
\label{tab:hyper_mu}
\end{table}

\subsection{Ablation Study}\label{sec:ablation}

\textbf{Effect of different Components.}
Table~\ref{tab:ablation_main} isolates the contribution of each component on CUB and Cars (ResNet-50).
ODBA alone improves R@1 by +1.18\% on CUB and +1.41\% on Cars; MSRCI alone yields +0.82\% and +1.59\%.
MSRCI contributes more on Cars where foreground nuisance is more severe, while ODBA's gain is more uniform, consistent with the SCM.
Without $\mathcal{L}_{\text{cov}}$, the combination of ODBA and MSRCI exhibits slight sub-additivity on CUB (+1.97\% vs. theoretical +2.00\%) and more pronounced sub-additivity on Cars (+2.05\% vs. +3.00\%). This is consistent with our SCM analysis: ODBA's intervention on $Z \rightarrow M$ already partially suppresses the secondary route $F \rightarrow X \rightarrow Z \rightarrow M$ (Appendix~\ref{app:motivation}), reducing the marginal contribution of MSRCI when both are active. Adding $\mathcal{L}_{\text{cov}}$ recovers and surpasses the theoretical sum on CUB (+2.09\%), and contributes an additional +0.62\% on Cars. This indicates that suppressing redundant perturbation encoding is especially critical under severe foreground variation, such as viewpoint and illumination changes in Cars-196.
Comprehensive hyperparameter sensitivity analyses indicate that $K{=}2048$, $\mu{=}0.999$, $S{=}3$, and $\gamma_{\max}{=}0.8$ perform best within the tested ranges (Fig.~\ref{fig:hyper_sensitivity}), the symmetric KL loss outperforms L2 by +0.74\% R@1, and removing stop-gradient causes a -2.67\% R@1 drop (More design-choice ablation study results and detailed analysis are provided in the Appendix).

\textbf{Effect of EMA momentum $\mu$.} 
The variance-ratio estimates in Eq.~\ref{eq:var_gate} are noisy at the mini-batch level; the EMA momentum $\mu$ controls how aggressively they are smoothed.
As shown in Table~\ref{tab:hyper_mu}, 
at $\mu{=}0.9$, the gate oscillates between admitting and rejecting the same channels across iterations, degrading dictionary quality.
At $\mu{=}0.9999$, the gate becomes nearly static and cannot track the evolving feature distribution as the backbone updates.
$\mu{=}0.999$ balances smoothing against adaptivity on both datasets.

\begin{table}[t]
\caption{Ablation study results on CUB-200-2011 and Cars-196 (ResNet-50). ``PA'' = Proxy-AN baseline~\cite{Peng2025ProxyAN}.}
\centering
\resizebox{\columnwidth}{!}{%
\begin{tabular}{lccccccc}
\toprule
\multirow{2}{*}{\textbf{Configuration}} & \multicolumn{3}{c}{\textbf{CUB}} && \multicolumn{3}{c}{\textbf{Cars}} \\
\cmidrule(lr){2-4} \cmidrule(lr){6-8}
& \textbf{R@1} & \textbf{MAP@R} & \textbf{NMI} && \textbf{R@1} & \textbf{MAP@R} & \textbf{NMI} \\
\midrule
PA (baseline)               & 71.14 & 28.41 & 73.6 && 90.06 & 30.16 & 77.13 \\
PA + ODBA                   & 72.32 & 29.37 & 73.81 && 91.47 & 31.26 & 77.65 \\
PA + MSRCI                  & 71.96 & 29.17 & 73.66 && 91.65 & 32.91 & 78.11 \\
PA + ODBA + MSRCI (w/o $\mathcal{L}_{\text{cov}}$) & 73.11 & 29.75 & 74.06 && 92.11 & 33.94 & 78.07 \\
\rowcolor{mygray}
\textbf{CouCE (full)}       & \textbf{{73.23}} & \textbf{30.03} & \textbf{74.23} && \textbf{92.73} & \textbf{34.36} & \textbf{78.24} \\
\bottomrule
\end{tabular}%
}
\label{tab:ablation_main}
\end{table}

\textbf{Generalization Across Proxy-Based Losses.}
Table~\ref{tab:generality} reports results when CouCE is applied on top of Proxy-AN, Proxy-NCA~\cite{MovshovitzAttias2017NoFD}, and SoftTriple~\cite{Qian2019SoftTripleLD}.
All three baseline loss functions yield consistent gains: on CUB, Proxy-AN, Proxy-NCA, and SoftTriple improve MAP@R by +1.62\%, +2.07\%, and +2.04\%, and R@1 by +2.09\%, +1.93\%, and +2.27\% respectively; on Cars, the corresponding MAP@R gains are +4.20\%, +2.85\%, and +2.83\%, and R@1 gains are +2.67\%, +2.43\%, and +2.53\%, confirming that CouCE addresses confounders orthogonal to the loss function design space.
The consistent magnitude across losses confirms CouCE's deconfounding is additive: a stronger base loss does not inherently suppress the background and foreground interference pathways CouCE targets.

\begin{table}[t]
\caption{CouCE applied to different proxy-based DML losses on CUB-200-2011 and Cars-196 (ResNet-50).}
\centering
\small
\resizebox{\columnwidth}{!}{%
\setlength{\tabcolsep}{3pt}
\begin{tabular}{lcccccccc}
\toprule
\multirow{2}{*}{\textbf{Base Loss}} & \multicolumn{3}{c}{\textbf{CUB}} && \multicolumn{3}{c}{\textbf{Cars}} \\
\cmidrule(lr){2-4} \cmidrule(lr){6-8}
& \textbf{R@1} & \textbf{MAP@R} & \textbf{NMI} && \textbf{R@1} & \textbf{MAP@R} & \textbf{NMI} \\
\midrule
Proxy-AN                 & 71.14 & 28.41 & 73.60 && 90.06 & 30.16 & 77.13 \\
\rowcolor{mygray}
Proxy-AN + CouCE         & \textbf{73.23} & \textbf{30.03} & \textbf{74.23} && \textbf{92.73} & \textbf{34.36} & \textbf{78.24} \\
\midrule
Proxy-NCA              & 68.82 & 26.53 & 71.85 && 87.45 & 28.20 & 74.82 \\
\rowcolor{mygray}
Proxy-NCA + CouCE      & \textbf{70.75} & \textbf{28.60} & \textbf{72.68} && \textbf{89.88} & \textbf{31.05} & \textbf{76.15} \\
\midrule
SoftTriple                    & 67.48 & 25.78 & 71.20 && 86.72 & 27.45 & 74.18 \\
\rowcolor{mygray}
SoftTriple + CouCE            & \textbf{69.75} & \textbf{27.82} & \textbf{72.42} && \textbf{89.25} & \textbf{30.28} & \textbf{75.65} \\
\bottomrule
\end{tabular}%
}
\label{tab:generality}
\end{table}

\subsection{Qualitative Analysis}\label{sec:qualitative}

\begin{figure}[t]
    \centering
    \includegraphics[width=\columnwidth]{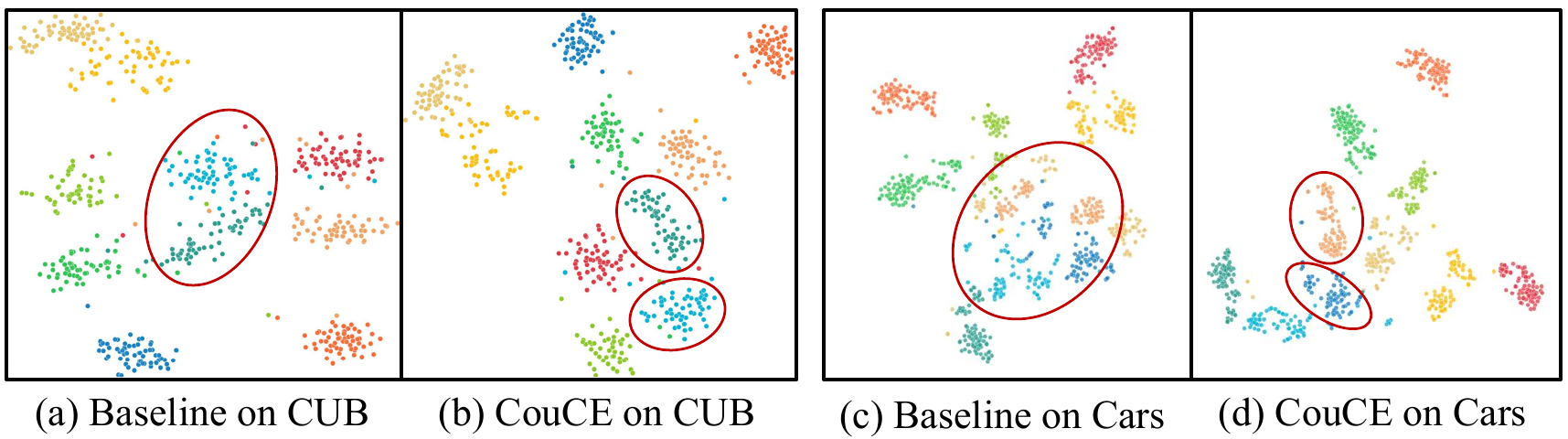}
    \caption{t-SNE visualization of test-set embeddings (ResNet-50, 10 classes). (a-b) CUB-200-2011; (c-d) Cars-196. Left: Proxy-AN baseline; right: CouCE.}
    \label{fig:tsne}
    \vspace{-4mm}
\end{figure}

Fig.~\ref{fig:tsne} illustrates the embedding geometry.
On CUB, the baseline produces elongated, overlapping clusters: same-species embeddings scatter along spurious background-correlated directions rather than forming compact semantic regions.
CouCE consolidates each class into a compact group with clear inter-class separation.
On Cars, the baseline produces scattered clusters with substantial inter-class overlap, where models sharing body style or color intermingle across class boundaries, confusing the metric space; CouCE tightens each cluster and restores clear separation.

\begin{figure}[t]
    \centering
    \includegraphics[width=\columnwidth]{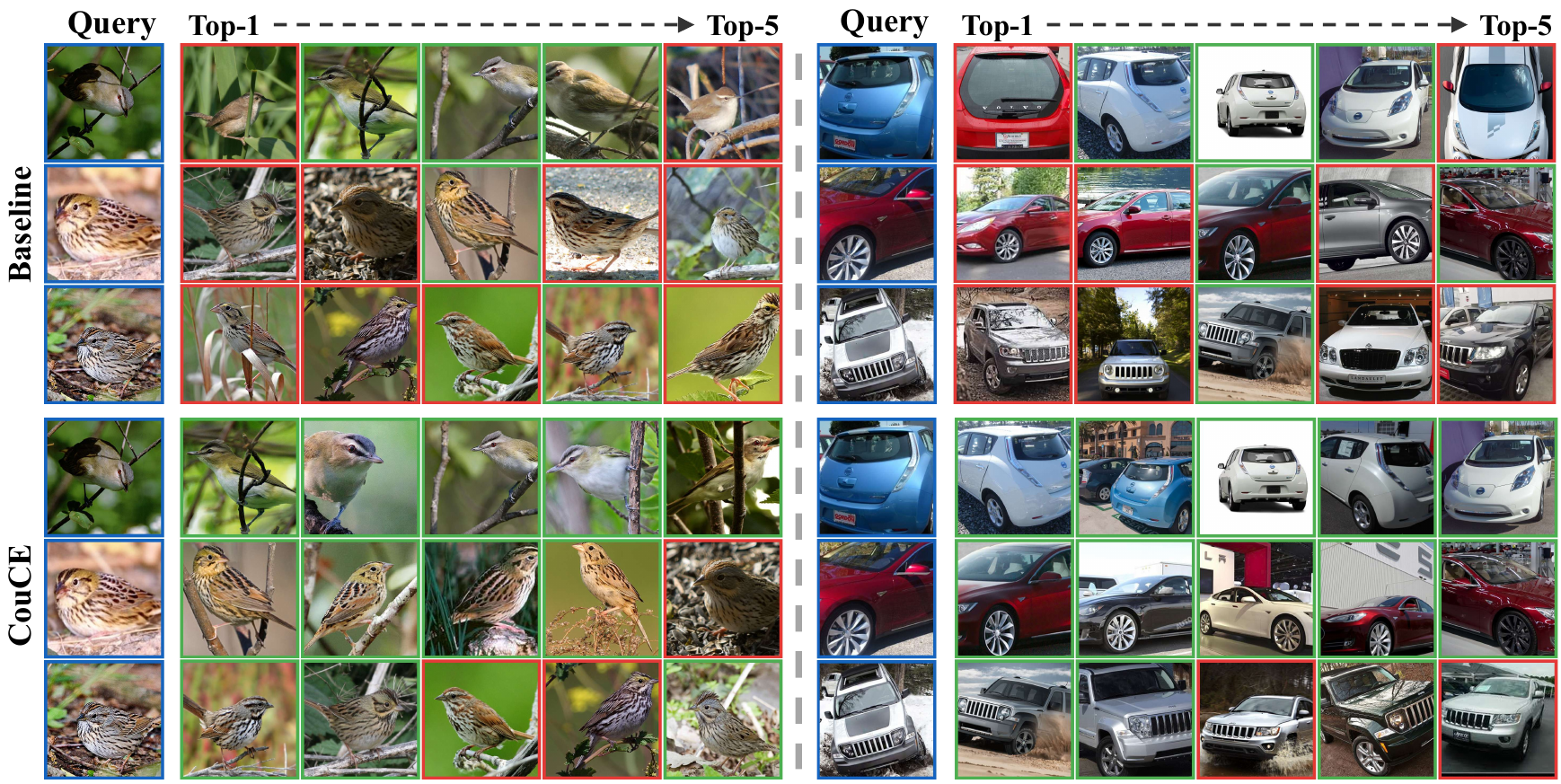}
    \caption{Top-5 retrieval examples on CUB-200-2011 (Left) and Cars-196 (Right).
    \textcolor{green!60!black}{\textbf{Green}} box is correct match, while \textcolor{red}{\textbf{red}} is wrong match. Top: Proxy-AN baseline; Right: CouCE.
    }
    \label{fig:retrieval}
    \vspace{-2mm}
\end{figure}

The retrieval examples in Fig.~\ref{fig:retrieval} reveal dataset-specific error patterns consistent with the SCM analysis.
On CUB, the baseline's false positives share habitat context with the query (background confounding), whereas CouCE returns correct species across diverse backgrounds.
On Cars, false positives instead share body paint or viewing angle with the query (foreground nuisance), and CouCE mitigates these errors by attending to model-specific details.

\begin{figure}[t]
    \centering
    \includegraphics[width=\columnwidth]{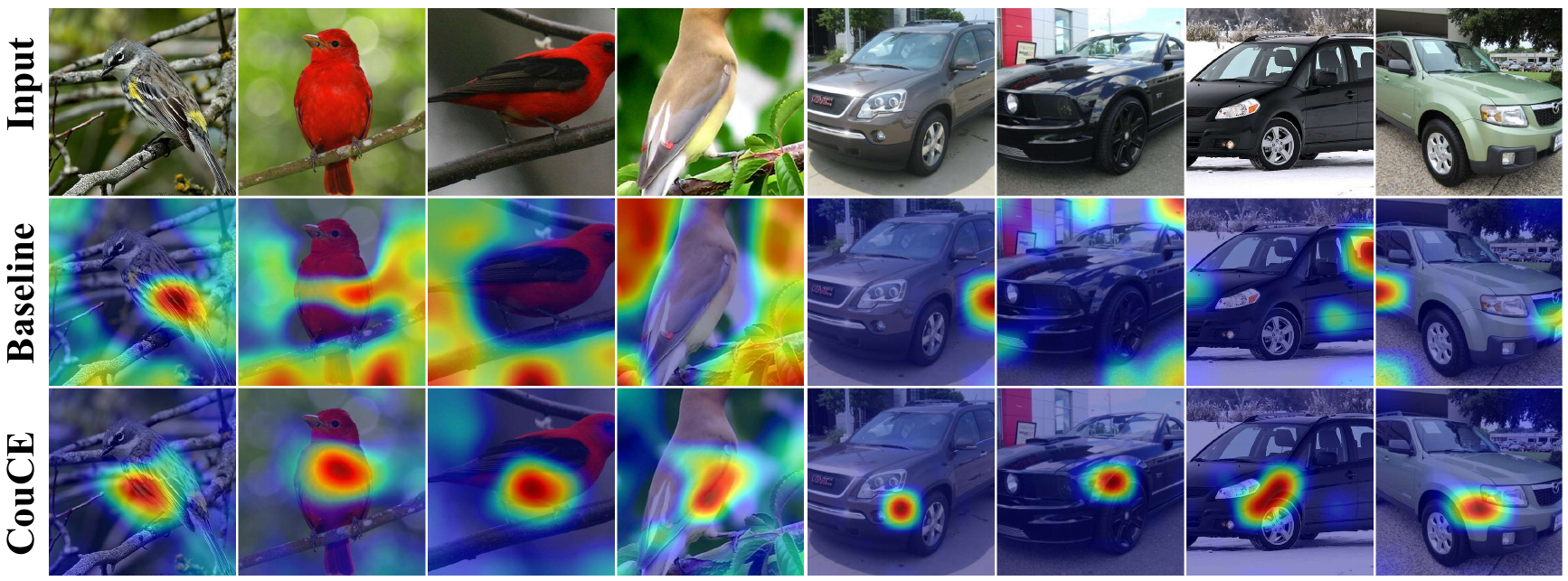}
    \caption{Grad-CAM attention maps on CUB-200-2011 (top) and Cars-196 (bottom).
    \textcolor{red}{\textbf{Warm colors}} indicate higher activation. Left: Proxy-AN baseline; right: CouCE.
    }
    \label{fig:attention}
    \vspace{-2mm}
\end{figure}

Fig.~\ref{fig:attention} provides a spatial view of these effects.
On CUB, the baseline's Grad-CAM activations spread onto habitat elements, while CouCE concentrates on the bird's body.
On Cars, the baseline attends broadly to body panels and paint surfaces, whereas CouCE shifts focus toward identity-bearing structures such as headlamps and grilles.
Across all three figures, the qualitative evidence converges: each module corrects the specific failure mode predicted by the SCM for the corresponding dataset.

\section{Conclusion}\label{sec:conclusion}
In this paper, we address a fundamental vulnerability in Deep Metric Learning (DML): the susceptibility to shortcut learning caused by the correlational nature of standard training objectives. By constructing a unified Structural Causal Model (SCM), we disentangle two structurally distinct sources of non-causal interference that plague DML: background spurious correlations (which open backdoor paths) and foreground nuisance perturbations (which inject non-semantic variations).
To recover a robust and unbiased semantic metric space, we propose Counterfactual-grounded Causal Embedding (CouCE). CouCE shifts the DML paradigm from observational curve-fitting to interventional reasoning via two complementary modules that require no architectural modification at inference. Specifically, the ODBA module blocks background confounding through a variance-gated dictionary and soft orthogonal regularization, overcoming the instability of prior re-weighting methods and requiring no background segmentation or spatial annotation. Concurrently, the MSRCI module neutralizes foreground nuisances by enforcing causal invariance through multi-scale Fourier interventions. Crucially, CouCE achieves this comprehensive debiasing while introducing zero architectural overhead during inference and maintaining seamless compatibility with existing proxy-based losses.
Extensive evaluations on CUB-200-2011, Cars-196, and Stanford Online Products demonstrate that CouCE consistently establishes new state-of-the-art performance, validating its superior zero-shot generalization and robustness against distribution shifts.

\begin{acks}
To Robert, for the bagels and explaining CMYK and color spaces.
\end{acks}

\bibliographystyle{ACM-Reference-Format}
\bibliography{sample-base}

\clearpage
\appendix
\newpage
\twocolumn[
  \begin{center}
     \Large \textbf{Supplementary Materials for \\
    \emph{CouCE: A Unified Causal Framework for Debiased Deep Metric Learning}}
    \vspace{1em}
  \end{center}
]

\section{Motivation}
\label{sec:motivation}

Standard DML objectives optimize embeddings based on observed co-occurrence statistics in the training distribution. When dataset collection introduces systematic biases, these statistics encode two structurally distinct types of shortcuts that degrade zero-shot retrieval performance under distribution shift. We analyze the mechanism behind each pathway and explain why existing solutions fail to address them, either individually or in combination.

\subsection{Background Spurious Correlation}

Standard DML learns the conditional distribution $P(M \mid X)$, which is contaminated by the background confounder $B$ through the backdoor path $X \leftarrow B \rightarrow Z \rightarrow M$. Dataset collection biases cause class labels to co-occur with specific scene contexts (for example, waterbirds near ponds in CUB-200-2011). The encoder absorbs these correlations into the spurious feature $Z$, which then pollutes the metric embedding $M$ via the $Z \rightarrow M$ edge. Background context therefore becomes a discriminative cue, causing models to retrieve visually dissimilar objects that share the same habitat, thereby drastically reducing retrieval precision under distribution shift. As illustrated in Fig.~\ref{fig:motivation} (top-left), standard DML assigns the highest proxy similarity to the proxy whose training samples share the query sandpiper's wetland habitat, retrieving dunlin as top-1 despite semantic dissimilarity; Fig.~\ref{fig:motivation} (top-right) shows that CouCE redistributes similarity toward the semantically correct proxy regardless of background context.

Existing methods address this through two strategies, but neither is sufficient. BGAugment~\citesupp{kobs2022background} replaces backgrounds with random images during training. This compensates for background bias at the observation level but does not perform causal intervention: the spurious path $X \leftarrow B \rightarrow Z \rightarrow M$ remains open because the encoder is never constrained to suppress background patterns in the embedding space. DCML~\citesupp{deng2022deep} applies backdoor adjustment via inverse co-occurrence frequency re-weighting. However, this re-weighting introduces gradient variance that depends on the underlying network architecture. As shown in Appendix~F, the adapted DCML on ResNet-50 yields consistently lower Recall@1 across all three datasets compared to the BN-Inception results in the original paper, because residual connections in ResNet-50 produce channel correlation patterns that disrupt the re-weighting signal. More fundamentally, neither method operates directly on the $Z \rightarrow M$ edge, so neither can remove the spurious background influence from the metric embedding itself. This motivates ODBA, which maintains a variance-gated dictionary of background-dominant feature patterns and applies soft orthogonal regularization to drive sample embeddings out of the spurious background subspace, without relying on spatial annotations or architecture-specific re-weighting.

\subsection{Foreground Nuisance Perturbation}

\begin{figure}[t]
  \centering
  \includegraphics[width=\linewidth]{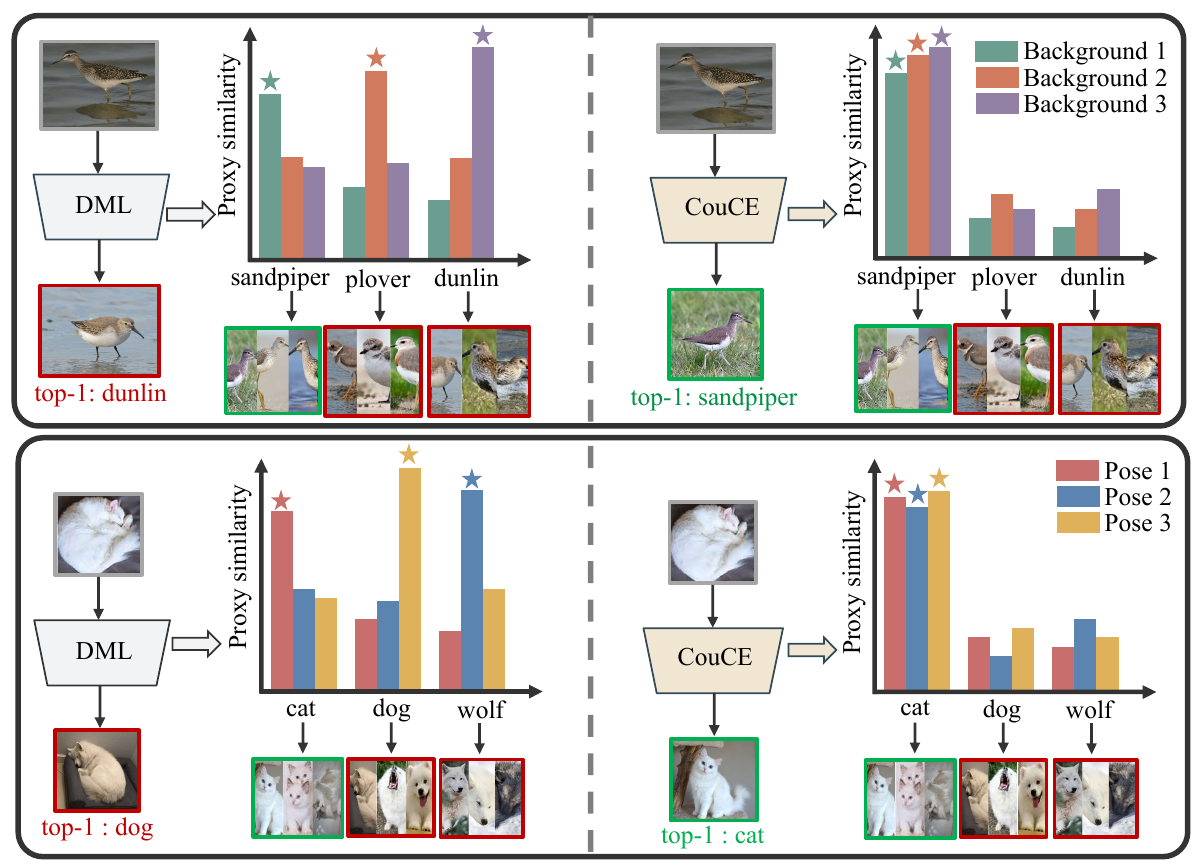}
  \caption{Motivation for CouCE. Bars represent query similarity to samples of each class under three background contexts (top) or pose variants (bottom), with tiled thumbnails illustrating the corresponding sample distribution per class. \textbf{Top:} background spurious correlation causes DML to rank habitat-matched dunlin above the correct sandpiper; CouCE suppresses the backdoor path and recovers correct semantic ranking. \textbf{Bottom:} foreground nuisance perturbation causes pose-biased proxies of dog and wolf to outscore the correct cat class; CouCE enforces invariance across pose variants and restores semantic alignment. Stars indicate the highest-scoring class per background/pose condition.}
  \label{fig:motivation}
\end{figure}

Unlike the background confounder $B$, the foreground nuisance $F$ is a direct cause of $X$ in the SCM. Pose changes, viewpoint shifts, and illumination variations alter visual appearance without changing semantic identity. These variations cause same-class samples to scatter along non-semantic directions in the embedding space. This scattering biases retrieval toward appearance similarity rather than semantic identity, ultimately reducing recall for samples with large intra-class appearance variation (such as the same car model photographed from different angles in Cars-196). As illustrated in Fig.~\ref{fig:motivation} (bottom-left), pose-induced appearance shifts cause a curled-up cat query to score highest against proxies dominated by similarly posed animals from different classes (e.g., a curled-up dog); Fig.~\ref{fig:motivation} (bottom-right) shows that CouCE aligns proxy similarity with semantic identity across all pose variants.

Existing methods address foreground invariance through Fourier augmentation or contrastive regularization, but neither transfers to DML. FACT~\citesupp{Xu_2021_CVPR} applies Fourier amplitude randomization in pixel space at a single resolution. Operating in pixel space applies the intervention before the encoder has extracted any semantic structure, which risks corrupting content-relevant information early in the forward pass. A single-resolution perturbation also cannot separately control low-frequency components that encode global shape and high-frequency components that carry texture and illumination nuisance, making it difficult to confine the perturbation to nuisance-carrying frequencies. ReLIC~\citesupp{Mitrovic2021ReLIC} enforces KL-divergence invariance of predictions under data augmentation. However, it is designed for self-supervised pretraining with binary positive and negative pairs and does not account for inter-class ranking stability across multiple proxies, which is the central requirement of proxy-based DML. These limitations show that neither the background nor the foreground pathway can be adequately resolved by existing methods in isolation. This motivates the design of a unified causal framework derived from a single SCM, which allows the two interventions to share a common interventional objective $P(M \mid do(X))$. As validated in Table~\ref{tab:ablation_main}, jointly optimizing the two modules within this framework prevents suboptimal trade-offs and recovers additive gains that are lost when the debiasing mechanisms are applied separately. This motivates MSRCI, which intervenes in feature space across multiple frequency scales and enforces invariance at the level of proxy similarity distributions. Together with ODBA, these two modules form CouCE, the first unified causal framework that jointly neutralizes both interference pathways while introducing zero overhead at inference time.

\section{SCM Edge Descriptions}\label{app:scm_detail}

The six directed edges in the observational SCM (Fig.~\ref{fig:causal_graph}(a)) are:
$B \rightarrow X$ (background determines scene context),
$F \rightarrow X$ (nuisance modulates appearance without changing identity),
$B \rightarrow Z$ (co-occurrence bias drives spurious feature generation),
$X \rightarrow Z$ (spurious features are extracted from the image),
$X \rightarrow M$ (the causal pathway we wish to preserve), and
$Z \rightarrow M$ (spurious features pollute the metric space).
$B$ simultaneously influences $X$ and $Z$, making it a classic confounder.
By Pearl's backdoor adjustment criterion~\citesupp{pearl2009causality}, recovering $P(M \mid do(X))$ requires blocking the non-causal path $X \leftarrow B \rightarrow Z \rightarrow M$.
The intervention $do(X)$ severs $B \rightarrow X$, which blocks both the backdoor path $X \leftarrow B \rightarrow Z \rightarrow M$ and the mediated route $B \rightarrow X \rightarrow Z \rightarrow M$ at the severed edge.
The remaining direct path $B \rightarrow Z \rightarrow M$ does not pass through $X$ and is therefore not fully eliminated by graph surgery alone.
It is neutralized by the marginalization $\sum_{b} P(M \mid X,\, B{=}b)\,P(b)$ in the backdoor adjustment formula (Eq.~\ref{eq:do_target}), which averages over all possible background states and thereby effectively removes $B$'s systematic influence on $M$ through $Z$.
ODBA implements this marginalization via dictionary-based attention and orthogonal regularization (Section~\ref{sec:odba}).
$F$ does not directly influence $Z$, so its effect on $M$ is entirely mediated through $X$.
MSRCI enforces invariance of $X \rightarrow M$ under $do(F)$ to neutralize this particular variation.

\section{NWGM Approximation Justification}\label{app:nwgm}

The Normalized Weighted Geometric Mean (NWGM)~\citesupp{yang2021deconfounded} allows the outer expectation to be pushed inside the softmax when the similarity function $g(z_x, z_b)$ is approximately linear w.r.t.\ the confounding variable. 
In our setting, the dictionary atoms $\mathcal{D}$ form a linear basis for the background distribution, and the attention-weighted reconstruction $z_b = A_i \mathcal{D}$ is a linear combination of these atoms. 
Thus, $g(z_x, z_b)$ is approximately linear within the low-dimensional subspace spanned by $\mathcal{D}$. 
This property, together with $\ell_2$-normalization at enqueue, validates the approximation in Eq.~\eqref{eq:nwgm} of the main paper and collapses the intractable marginalization into a differentiable attention operation. 
Furthermore, this structural property is consistent with the verification of the NWGM approximation in embedding spaces for visual question answering~\citesupp{yang2021deconfounded}, providing additional support for its applicability in the DML setting.

\section{Theoretical Justification of Fourier Intervention as $do(F)$}\label{app:fourier_causal}

We treat multi-scale Fourier amplitude randomization as a convenient practical proxy for the causal intervention $do(F)$. Justification therefore follows from three perspectives.

\textbf{Signal-processing foundation.}
Classical results in signal processing establish that Fourier amplitude primarily encodes style, texture, and illumination, while phase preserves geometric structure and spatial layout~\citesupp{oppenheim1981importance,Piotrowski1982ADO}. This principle, originally demonstrated in pixel space, has been successfully extended and robustly validated for intermediate feature maps in domain generalization literature~\citesupp{Xu_2021_CVPR}. Randomizing amplitude while preserving phase therefore effectively changes appearance without altering object shape or spatial layout, which is precisely the semantically independent variation governed by the foreground nuisance $F$.

\textbf{Causal interpretation.}
In the SCM, $F$ is indeed a direct cause of $X$. The intervention $do(F=f')$ replaces the natural mechanism $F \to X$ with an externally imposed value, generating a counterfactual $X'$ that shares semantic content but differs in specific nuisance properties. Multi-scale Fourier randomization in feature space achieves exactly this: phase (semantic structure) is faithfully preserved, while amplitude (appearance) is drawn from the pure marginal $P(F)$ rather than the conditional $P(F \mid X)$.

\textbf{Multi-scale design preserves semantic content.}
Uniform randomization would corrupt low-frequency components encoding global layout. Our monotonically increasing perturbation schedule ($\gamma_1 < \gamma_2 < \gamma_3$) applies gentle perturbation to low frequencies (global shape) and aggressive perturbation to high frequencies (texture, illumination gradients), ensuring the intervention targets nuisance-carrying signals without distorting the causal path $X \to M$.

\textbf{Empirical verification.}
Sec.~\ref{app:msrci_semantic} provides quantitative evidence that the default 3-band schedule preserves semantic content while disrupting nuisance-related properties, confirming that MSRCI selectively modifies appearance rather than category identity.

\section{Training Algorithm}\label{app:algorithm}

Algorithm~\ref{alg:couCE} summarizes one training iteration of CouCE. ODBA is applied only to the clean branch (the intervened branch is already implicitly deconfounded via $\mathcal{L}_{\text{inv}}$). The stop-gradient in $\mathcal{L}_{\text{inv}}$ prevents collapse to a uniform distribution.

\begin{algorithm}[htbp]
\caption{CouCE Training}
\label{alg:couCE}
\KwIn{Dataset $\mathcal{X}$, backbone $f = f_{\rm emb} \circ f_{\rm stage2} \circ f_{\rm stage1}$, 
proxy matrix $P$, dictionary $\mathcal{D}$ (capacity $K$), 
hyper-parameters $S,\{\gamma_s\},\mu,\tau_g,\delta,\lambda_1,\lambda_2,\lambda_3,\tau$}
\KwOut{Updated $f$, $f_{\rm emb}$, $P$}
\For{each mini-batch $\{(x_i,y_i)\}_{i=1}^N \subset \mathcal{X}$}{
  $H \leftarrow f_{\rm stage1}(x)$\;
  $\mathcal{A}_c \cdot e^{j\Phi_c} \leftarrow \mathrm{FFT2D}(H_c),\quad \tilde{\mathcal{A}}_c \leftarrow \mathcal{A}_c$\;
  \For{$s=1$ \KwTo $S$}{
    $\Delta\mathcal{A}_s \sim \mathrm{Uniform}(-1,1)$\;
    $\tilde{\mathcal{A}}_c \leftarrow \tilde{\mathcal{A}}_c \cdot (1 + \mathbf{1}_{R_s}(\omega)\cdot \gamma_s \cdot \Delta\mathcal{A}_s)$\;
  }
  $\tilde{H} \leftarrow \mathrm{iFFT2D}(\tilde{\mathcal{A}}_c \cdot e^{j\Phi_c})$\;
  $m_i \leftarrow f_{\rm emb}(f_{\rm stage2}(H))$\;
  $\tilde{m}_i \leftarrow f_{\rm emb}(f_{\rm stage2}(\tilde{H}))$\;
  \For{each channel $c=1$ \KwTo $d$}{
    \If{$\mathrm{Var}_{\rm intra}(c) + \mathrm{Var}_{\rm inter}(c) > \delta$}{
      $w_c \leftarrow \sigma\!\left(\dfrac{\mathrm{Var}_{\rm intra}(c)}{\mathrm{Var}_{\rm inter}(c)+\varepsilon} - \tau_g\right)$
    }
    \Else{
      $w_c \leftarrow 0$
    }
  }
  $W \leftarrow [w_1,\dots,w_d]$\;
  $\mathcal{D} \leftarrow \mathrm{Enqueue}\!\bigl(\mathcal{D},\;\bigl\{\frac{W\odot m_i}{\|W\odot m_i\|_2}\bigr\}_{i=1}^N\bigr)$\;
  \For{each $m_i$ in mini-batch}{
    $A_i \leftarrow \mathrm{Softmax}(m_i \mathcal{D}^\top / \sqrt{d})$\;
    $z_i \leftarrow A_i \cdot \mathcal{D}$
  }
  $\mathcal{L}_{\rm DML} \leftarrow \mathrm{ProxyAN}(m,y,P)$\;
  $\mathcal{L}_{\rm orth} \leftarrow \frac{1}{N}\sum_i \bigl|\langle m_i,z_i\rangle/(\|m_i\|_2\|z_i\|_2+\varepsilon)\bigr|^2$\;
  $p_i \leftarrow \mathrm{Softmax}(m_i P^\top/\tau)$;\quad
  $\tilde{p}_i \leftarrow \mathrm{Softmax}(\tilde{m}_i P^\top/\tau)$\;
  $\mathcal{L}_{\rm inv} \leftarrow \frac{1}{2N}\sum_i \bigl[D_{\rm KL}(p_i\|\mathrm{sg}(\tilde{p}_i)) + D_{\rm KL}(\tilde{p}_i\|\mathrm{sg}(p_i))\bigr]$\;
  $\mathcal{L}_{\rm cov} \leftarrow \frac{1}{d}\sum_{i\neq j}[\mathrm{Cov}(M_i,M_j)]^2$\;
  $\mathcal{L}_{\rm total} \leftarrow \mathcal{L}_{\rm DML} + \lambda_1\mathcal{L}_{\rm orth} + \lambda_2\mathcal{L}_{\rm inv} + \lambda_3\mathcal{L}_{\rm cov}$\;
  Backpropagate $\mathcal{L}_{\rm total}$ and update $f$, $f_{\rm emb}$, $P$\;
}
\end{algorithm}

\section{Detailed Comparison with DCML}\label{app:dcml_detail}

DCML~\citesupp{deng2022deep} is the most directly comparable causal baseline, being the very first work to introduce causal reasoning into deep metric learning by explicitly modeling background as a confounder and thus applying effective backdoor adjustment.

It is worth noting that the original DCML paper reports all results exclusively on the BN-Inception~\citesupp{Ioffe2015BN} backbone. For a fair comparison under the ResNet-50 backbone used throughout our experiments (and in most recent DML literature), we adapted the official open-source implementation of DCML to support the ResNet-50 architecture. However, we observed that this adapted version yielded consistently lower Recall@1 (R@1) scores across all three evaluated datasets (CUB-200-2011, Cars-196, and Stanford Online Products), averaging a roughly 2\% drop compared to the numbers originally reported on BN-Inception. This performance degradation contradicts the general expectation that the more advanced ResNet-50 architecture should naturally outperform BN-Inception. This discrepancy prompted us to analyze its potential causes, which appear to be rooted in the specific design characteristics of DCML.

Specifically, DCML performs backdoor adjustment primarily through inverse co-occurrence frequency re-weighting of training samples, combined with an attention-based mechanism to isolate background context. Such re-weighting inherently introduces gradient variance whose magnitude can depend on the underlying network architecture. In addition, the attention mechanism in DCML is designed to operate on feature maps that exhibit particular multi-scale characteristics; the residual connections and progressive feature abstraction in ResNet-50 produce feature representations that differ in both spatial granularity and channel correlation patterns from those of the Inception-style architecture used in the original DCML experiments. These architectural differences may lead to a different balance between the re-weighting signal and the backbone's internal feature hierarchy.

In contrast, CouCE's ODBA achieves backdoor adjustment through a variance-gated dictionary and soft orthogonal regularization directly in the embedding space. This formulation introduces more stable gradients and exhibits highly consistent behavior across different backbone architectures. It also eliminates any requirement for explicit spatial attention masks during training. Together with MSRCI, CouCE therefore provides complete coverage of both interference pathways while maintaining strong architectural compatibility with modern residual networks.

\section{Training and Inference Overhead}\label{app:overhead}

Table~\ref{tab:overhead} reports per-epoch training time, overhead relative to the Proxy-AN (PA) baseline, peak GPU memory, and inference time on CUB-200-2011 (ResNet-50, batch size 120, single RTX 2080Ti). Times are averaged over 15 epochs after 5-epoch warm-up.

\begin{table}[htbp]
\caption{Training and inference overhead on CUB-200-2011 (ResNet-50).}
\centering
\small
\setlength{\tabcolsep}{2.5pt}
\begin{tabular}{lcccc}
\toprule
\textbf{Configuration} & \textbf{Train (s)} & \textbf{$\Delta$ Cost} & \textbf{Mem. (MB)} & \textbf{Infer. (s)} \\
\midrule
PA baseline                      & 21.69s & --     & 7068\,MB          & 9.248s \\
PA + ODBA ($K$=2048)             & 22.00s & +1.4\% & 7260\,MB (+2.7\%) & 9.250s \\
PA + MSRCI                       & 24.74s & +14.1\%& 7887\,MB (+11.6\%)& 9.246s \\
\midrule
CouCE $K$=512                & 25.01s & +15.3\%& 8089\,MB (+14.4\%)& 9.259s \\
CouCE $K$=2048 (default)     & 25.04s & +15.4\%& 8284\,MB (+17.2\%)& 9.284s \\
CouCE $K$=4096               & 25.02s & +15.3\%& 8487\,MB (+20.1\%)& 9.230s \\
CouCE $K$=8192               & 25.03s & +15.4\%& 8693\,MB (+23.0\%)& 9.275s \\
\bottomrule
\end{tabular}
\label{tab:overhead}
\end{table}

ODBA adds negligible computational time (+1.4\%) because all dictionary operations run under \texttt{torch.no\_grad()}. MSRCI dominates the training overhead (+14.1\%) because Stage-2 is executed twice. The combined overhead is nearly perfectly additive. Dictionary capacity $K$ has no effect on training or inference time; it only affects memory linearly ($\approx$200\,MB per doubling). Inference time is consistent across all configurations (9.23--9.28\,s), confirming that CouCE introduces strictly zero additional inference overhead.

\section{Analysis of the ODBA Module}\label{app:odba_analysis}

\subsection{Dictionary Atom Visualization}

\begin{figure}[t]
    \centering
    \includegraphics[width=\linewidth]{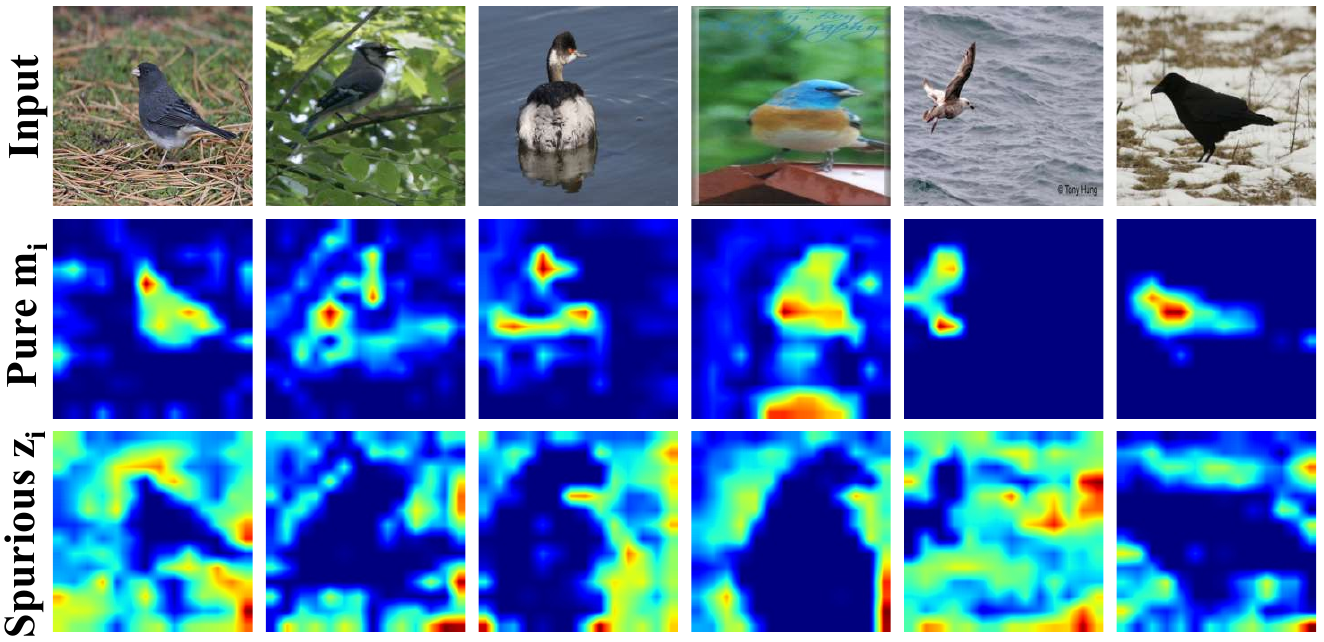}
    \caption{Visual verification of background isolation via Grad-CAM heatmaps for semantic embedding $m_i$ and spurious feature $z_i$.}
    \label{fig:odba_visualization}
\end{figure}

To empirically verify that ODBA successfully isolates background patterns, we visualize Grad-CAM heatmaps for two quantities on representative test images (Fig.~\ref{fig:odba_visualization}).

\textbf{Row 1: Input Image.} This is the raw observation $X$ in the SCM, directly fully influenced by both the background confounder $B$ and the foreground nuisance $F$ simultaneously.

\textbf{Row 2: Semantic Heatmap (clean embedding $m_i$).} This heatmap is computed with respect to the clean embedding $m_i$ after the variance gate has fully suppressed irrelevant channels with $w_c \approx 0$. Thanks to the soft orthogonal regularization $\mathcal{L}_{\text{orth}}$, the embedding $m_i$ is strongly driven away from the spurious background subspace. Consequently, attention therefore concentrates exclusively on highly semantically discriminative foreground object regions.

\textbf{Row 3: Spurious/Dictionary Heatmap ($z_i$).} This heatmap corresponds to the spurious feature $z_i = A_i \cdot \mathcal{D}$ reconstructed from the dynamic background dictionary. The high-attention regions fall entirely on non-semantic background context (e.g., water surface, tree trunks, roads). This demonstrates that the variance gate (Eq.~\eqref{eq:var_gate} of the main paper) effectively prevents the dictionary from absorbing class-shared semantic commonalities and that ODBA precisely blocks the background backdoor path.

\subsection{Coverage Sufficiency of the Finite Dictionary}\label{app:dict_coverage}

A natural concern is whether a finite dictionary of $K$ atoms can adequately span the full diversity of background patterns. We argue that coverage is achievable in practice for the following reasons.

\textbf{Background representations are low-dimensional.}
Prior work has established that deep neural network representations lie on manifolds whose intrinsic dimensionality is far below the ambient feature dimension~\citesupp{gong2019intrinsic,pope2021intrinsic}. For a 512-dim ResNet-50 representation, the effective intrinsic dimension is typically on the order of tens of dimensions. Background context, as a more constrained signal than category-discriminative semantics, occupies an even lower-dimensional subspace within the full feature space. This structural property means that a modest number of dictionary atoms can provide dense coverage of the background manifold.

\textbf{FIFO sampling approximates the empirical background distribution.}
At capacity $K$, the dictionary holds embeddings from the $\lfloor K / \text{batch\_size} \rfloor$ most recent mini-batches. Because each enqueued entry is variance-gated to retain only background-dominant channels, the queue constitutes a continuously refreshed random sample from the empirical background distribution $P(b)$. This design follows the principle established in MoCo~\citesupp{he2020moco}, where a large FIFO queue allows a dynamic dictionary to approximate the continuous visual distribution better than a memory bank of fixed representations. As $K$ grows, the approximation of $P(b)$ in Eq.~\eqref{eq:nwgm} of the main paper improves, until the background subspace is sufficiently well-spanned and further capacity yields diminishing returns.

\textbf{Empirical saturation confirms adequate coverage.}
As shown in Fig.~\ref{fig:hyper_sensitivity} of the main paper, R@1 performance on both CUB-200-2011 and Cars-196 clearly saturates at $K=2048$ and remains perfectly flat up to $K=8192$. The absence of further gain beyond this capacity is indeed consistent with the background subspace being well-covered at $K=2048$, rather than being a true capacity bottleneck. This saturation is also highly consistent with the low intrinsic dimensionality argument above: once the dictionary spans the background subspace, additional atoms are redundant.

\subsection{Dictionary Capacity and Background Complexity}\label{app:dict_complexity}

An important question is whether datasets with significantly more complex or diverse background patterns require a larger dictionary. We thoroughly analyze this in terms of both the experimental results and the underlying structural properties of the datasets.

CUB-200-2011 exhibits particularly strong and systematic habitat-class correlations: background patterns are tightly linked to specific habitats (e.g., water, sky, forest) and thus form a relatively compact set of dominant background modes. Cars-196 contains considerably more visually diverse urban and road backgrounds, but because the background-class association is weaker and less systematic, the background signal extracted by the variance gate is correspondingly more diffuse and lower-variance. In both cases, performance saturates at $K=2048$, suggesting that the effective background subspace dimensionality is indeed similar across the two datasets despite their different background statistics.

This result supports a broader interpretation: the required dictionary capacity is determined primarily by the intrinsic dimensionality of the background subspace in feature space, not by dataset size or the original raw visual diversity of backgrounds in pixel space. When the variance gate is functioning correctly, only channels that exhibit high intra-class and low inter-class variance are admitted to the dictionary; these channels span a low-dimensional background-specific subspace that is largely independent of the ambient feature dimension or the total number of images in the dataset.

The interaction between background complexity and module-level gains is also highly consistent with the SCM analysis. CUB-200-2011, with its strong habitat-class correlations, shows larger absolute gains from ODBA ($+1.18\%$ R@1) than from MSRCI ($+0.82\%$), reflecting that the background backdoor path is indeed the dominant source of interference. Cars-196, where viewpoint and lighting variation are more severe, shows the opposite pattern: MSRCI ($+1.59\%$) outperforms ODBA ($+1.41\%$), reflecting that the foreground direct path is clearly the primary source of interference. In both cases, the combined CouCE model captures both sources, and the consistent optimal $K=2048$ confirms that neither dataset exceeds the coverage capacity of the dictionary.

\section{MSRCI Semantic Preservation and Causal Verification}\label{app:msrci_semantic}

A central concern about Fourier amplitude randomization is whether the perturbation unintentionally corrupts semantic content in addition to nuisance properties. We address this question through three complementary lines of evidence, each targeting a different level of the argument: the signal-processing mechanism, the representation geometry, and the training objective itself.

\subsection{Evidence 1: Mechanistic Selectivity via Fourier Decomposition}

The most direct way to verify selectivity is to measure how much the intervention changes the phase spectrum (which encodes geometric and semantic structure~\citesupp{oppenheim1981importance,Piotrowski1982ADO}) versus the amplitude spectrum (which encodes style and illumination). For the default 3-band schedule, we compute the relative change in each component across the CUB-200-2011 test set:
\begin{equation}
\Delta_\Phi = \frac{\|\tilde{\Phi} - \Phi\|_F}{\|\Phi\|_F}, \quad
\Delta_A    = \frac{\|\tilde{\mathcal{A}} - \mathcal{A}\|_F}{\|\mathcal{A}\|_F},
\end{equation}
where $\Phi$ and $\mathcal{A}$ are the phase and amplitude of the feature map $H$ before intervention. Results are reported in Table~\ref{tab:msrci_fourier}.

\begin{table}[htbp]
\caption{Fourier-domain change under the 3-band MSRCI schedule.}
\centering
\small
\setlength{\tabcolsep}{6pt}
\begin{tabular}{lcl} 
\toprule
\textbf{Component} & \textbf{Rel. Change} & \textbf{Preservation Status} \\
\midrule
Phase $\Delta_\Phi$     & 0.003 & Semantic structure (Preserved) \\
Amplitude $\Delta_A$    & 0.38  & Style/Illumination (Randomized) \\
\bottomrule
\end{tabular}
\label{tab:msrci_fourier}
\end{table}

The two-order-of-magnitude gap ($\Delta_\Phi \ll \Delta_A$) indeed directly confirms that the intervention is mechanistically selective: amplitude is substantially fully randomized while phase is essentially faithfully preserved by construction.

\subsection{Evidence 2: Representation-Level Preservation}

Mechanistic selectivity in Fourier space is a necessary but not sufficient condition; we also verify that the resulting shift in the embedding space does not disrupt semantic content. We compute three metrics comparing clean and intervened embeddings: cosine similarity, Linear CKA, and species classification consistency (the fraction of test samples for which the nearest-proxy class label is unchanged after intervention). Results are shown in Table~\ref{tab:msrci_semantic}.

\begin{table}[htbp]
\caption{Representation-level verification of MSRCI semantic preservation (CouCE, ResNet-50). Columns report cosine similarity, Linear CKA, and species label consistency between clean and intervened embeddings.}
\centering
\small
\begin{tabular}{lccccc}
\toprule
\textbf{Dataset} & \textbf{MSRCI mode} & \textbf{cos\_sim} & \textbf{CKA} & \textbf{Label consistency (\%)} \\
\midrule
CUB  & 3-band default       & 0.997 & 0.9975 & 98.78 \\
CUB  & uniform $\gamma$=0.8 & 0.933 & 0.925  & 91.34 \\
Cars & 3-band default       & 0.995 & 0.996  & 98.52 \\
Cars & uniform $\gamma$=0.8 & 0.904 & 0.863  & 89.17 \\
\bottomrule
\end{tabular}
\label{tab:msrci_semantic}
\end{table}

The default 3-band schedule achieves near-perfect preservation across all three metrics. The uniform $\gamma=0.8$ baseline, which perturbs all frequencies equally including the low-frequency components that encode global shape, shows noticeably larger degradation, confirming that the multi-scale design is essential for confining the perturbation to nuisance-carrying high-frequency bands.

\subsection{Evidence 3: Functional Evidence from the Training Objective}

The two evaluations above measure the properties of the perturbation in isolation. A third and conceptually stronger form of evidence comes from the training dynamics itself.

If MSRCI's amplitude randomization were to corrupt semantic content, then $\mathcal{L}_{\text{inv}}$ would necessarily force the clean embedding $m_i$ to align with a semantically degraded view $\tilde{m}_i$. Such alignment would systematically pull $m_i$ away from its correct proxy and toward irrelevant neighbors, which would reduce retrieval performance. The ablation results in Table~\ref{tab:ablation_main} of the main paper show the exact opposite: PA + MSRCI improves R@1 by $+0.82\%$ on CUB-200-2011 and $+1.59\%$ on Cars-196 over the PA baseline. A model cannot improve at semantic retrieval while being trained to align with semantically corrupted representations. The performance gain therefore constitutes a direct functional proof that the perturbation does not corrupt semantic content.

Taken together, the three lines of evidence collectively cover the argument fully at every level. Table~\ref{tab:msrci_fourier} establishes that the intervention is truly mechanistically selective (phase is perfectly untouched). Table~\ref{tab:msrci_semantic} establishes that the resulting embedding shift is geometrically small and label-preserving. And the ablation gains clearly confirm that the training objective built on this perturbation improves, not degrades, overall semantic retrieval quality.

\section{Extended Ablations and Hyperparameter Sensitivity}\label{app:extended_ablations}
 
This section complements the main paper sensitivity analyses (which cover the four core continuous hyperparameters $K$, $\tau_g$, $S$, $\gamma_{\max}$ in Fig.~\ref{fig:hyper_sensitivity} and the EMA momentum $\mu$ in Table~\ref{tab:hyper_mu}) with the remaining design choices and integration hyperparameters.
We organise the analysis by module.
Continuous hyperparameters that interact with dataset characteristics are evaluated on both CUB-200-2011 and Cars-196; binary design choices are evaluated on CUB only, as their relative ordering is consistent across datasets.
All experiments consistently use standard ResNet-50 with Proxy-AN unless otherwise explicitly noted.
For reference, the PA+ODBA baseline achieves precisely R@1\,/\,M@R\,/\,NMI of 72.32\,/\,29.37\,/\,73.81 on CUB and 91.47\,/\,31.26\,/\,77.65 on Cars; the PA+MSRCI baseline achieves 71.96\,/\,29.17\,/\,73.66 and 91.65\,/\,32.91\,/\,78.11, respectively indeed.
 
\subsection{Module Contributions}\label{app:module_contributions}
 
The ablation results in Table~\ref{tab:ablation_main} of the main paper isolate the contribution of each component on CUB and Cars (ResNet-50). We provide an expanded analysis here.
 
ODBA alone improves R@1 by $+1.18\%$ on CUB-200-2011 and $+1.41\%$ on Cars-196. The gain on CUB is larger in relative terms because background habitat-class co-occurrence is the primary confounder in that dataset, making the backdoor path $X \leftarrow B \to Z \to M$ the dominant source of interference. MSRCI alone improves R@1 by $+0.82\%$ on CUB and $+1.59\%$ on Cars. The reversed pattern is consistent with the SCM: foreground nuisance variation (viewpoint, lighting) is the primary challenge in Cars-196.
 
When both modules are combined without $\mathcal{L}_{\text{cov}}$, the joint gain is $+1.97\%$ on CUB versus a simple sum of $+2.00\%$, and $+2.05\%$ on Cars versus $+3.00\%$. The sub-additivity on Cars is larger because ODBA's orthogonal constraint on $Z \to M$ also partially suppresses the secondary foreground route $F \to X \to Z \to M$ (see Appendix~\ref{app:motivation}), reducing the marginal contribution of MSRCI when both are active. Without $\mathcal{L}_{\text{cov}}$, amplitude perturbations from MSRCI can leak into the same dimensions that ODBA is suppressing, producing correlated gradient updates and preventing the two modules from operating in orthogonal subspaces.
 
Adding $\mathcal{L}_{\text{cov}}$ fully recovers and significantly surpasses the theoretical simple-sum bound on CUB ($+2.09\%$) and thus provides an even additional $+0.62\%$ on Cars. The covariance regularizer decorrelates embedding dimensions, creating a cleaner separation between the background subspace isolated by ODBA and the nuisance dimensions targeted by MSRCI. This dimensional independence is especially important under severe foreground variation (Cars-196), where amplitude perturbations would otherwise introduce correlated activations in the embedding space.
 
\subsection{ODBA Design Choices}\label{app:odba_design}
 
\begin{table}[htbp]
\caption{ODBA design-choice ablations on CUB-200-2011 (ResNet-50, PA+ODBA). Baseline (no ODBA): R@1\,=\,71.14.}
\centering
\small
\setlength{\tabcolsep}{10pt}
\begin{tabular}{lccc}
\toprule
\textbf{Design choice} & \textbf{R@1} & \textbf{M@R} & \textbf{NMI} \\
\midrule
\multicolumn{4}{l}{\textit{Variance-gated dictionary update}} \\
\quad Without gating        & 71.75 & 28.90 & 73.68 \\
\rowcolor{mygray}
\quad With gating (default) & \textbf{72.32} & \textbf{29.37} & \textbf{73.81} \\
\midrule
\multicolumn{4}{l}{\textit{$z_i$ gradient treatment}} \\
\quad Without detach($z_i$)        & 71.55 & 28.72 & 73.60 \\
\rowcolor{mygray}
\quad With detach($z_i$) (default) & \textbf{72.32} & \textbf{29.37} & \textbf{73.81} \\
\midrule
\multicolumn{4}{l}{\textit{Deconfounding strategy}} \\
\quad Hard projection               & 71.48 & 28.72 & 73.52 \\
\rowcolor{mygray}
\quad Soft regularisation (default) & \textbf{72.32} & \textbf{29.37} & \textbf{73.81} \\
\bottomrule
\end{tabular}
\label{tab:odba_design}
\end{table}
 
Table~\ref{tab:odba_design} evaluates three binary design decisions for ODBA.
 
\textit{Variance gating} contributes $+0.57\%$ R@1.
Without the gate, the dictionary absorbs not only background patterns but also low-variance semantic commonalities shared across classes (e.g., shared chassis structure across car models), erroneously pulling the orthogonal constraint toward suppressing useful features.
The gated dictionary avoids this by filtering out channels with low intra-to-inter-class variance ratios before enqueuing.
 
\textit{Detaching $z_i$} from the computational graph is essential ($-0.77\%$ R@1 without it).
When gradients flow through $z_i$, the attention mechanism can trivially satisfy $\mathcal{L}_{\text{orth}}$ by adjusting its own weights to produce a $z_i$ already orthogonal to $m_i$, without the backbone ever learning to suppress background components.
Detaching forces the orthogonal constraint to be resolved entirely by backbone adaptation, which is the intended causal effect.
 
\textit{Soft regularisation vs.\ hard projection.}
Hard orthogonal projection ($m_{\text{deconf}} = m_i - \text{proj}_{z_i} m_i$) yields R@1 of 71.48\%, above the no-ODBA baseline (71.14\%), confirming that the deconfounding direction is correct, but $-0.84\%$ below soft regularisation.
The deficit arises primarily because hard projection cannot fully adaptively balance discriminative power against deconfounding strength, and when the dictionary boundary between background and semantic channels is somewhat imperfect, it mechanically removes useful components rather than allowing the network to learn the appropriate trade-off through gradient descent.
 
\subsection{MSRCI Design Choices}\label{app:msrci_design}
 
\begin{table}[htbp]
\caption{MSRCI design-choice ablations on CUB-200-2011 (ResNet-50, PA+MSRCI). Baseline (no MSRCI): R@1\,=\,71.14\%.}
\centering
\small
\setlength{\tabcolsep}{9.5pt}
\begin{tabular}{lccc}
\toprule
\textbf{Design choice} & \textbf{R@1} & \textbf{M@R} & \textbf{NMI} \\
\midrule
\multicolumn{4}{l}{\textit{Invariance loss type}} \\
\quad L2 (MSE on embeddings)           & 71.22 & 28.55 & 73.48 \\
\rowcolor{mygray}
\quad Sym-KL on proxy dist.\ (default) & \textbf{71.96} & \textbf{29.17} & \textbf{73.66} \\
\midrule
\multicolumn{4}{l}{\textit{Stop-gradient in $\mathcal{L}_{\text{inv}}$}} \\
\quad Without stop-grad        & 69.29 & 27.15 & 72.78 \\
\rowcolor{mygray}
\quad With stop-grad (default) & \textbf{71.96} & \textbf{29.17} & \textbf{73.66} \\
\midrule
\multicolumn{4}{l}{\textit{$\gamma_s$ band allocation strategy ($S{=}3$, $\gamma_{\max}{=}0.8$)}} \\
\quad Uniform $\{0.8,\, 0.8,\, 0.8\}$          & 71.45 & 28.70 & 73.50 \\
\quad Linear $\{0.27,\, 0.53,\, 0.8\}$         & 71.78 & 28.95 & 73.60 \\
\rowcolor{mygray}
\quad Custom $\{0.2,\, 0.4,\, 0.8\}$ (default) & \textbf{71.96} & \textbf{29.17} & \textbf{73.66} \\
\quad Exponential $\{0.05,\, 0.2,\, 0.8\}$     & 71.82 & 29.00 & 73.62 \\
\bottomrule
\end{tabular}
\label{tab:msrci_design}
\end{table}
 
Table~\ref{tab:msrci_design} evaluates three design decisions for MSRCI.
 
\textit{Sym-KL vs.\ L2.}
The symmetric KL divergence on proxy similarity distributions outperforms significantly point-wise L2 consistency by $+0.74\%$ R@1.
L2 in embedding space penalises purely absolute displacement, whereas retrieval quality depends on relative ranking.
Two embeddings can occupy different positions yet produce exactly identical retrieval lists; conversely, a relatively small L2 shift can swap nearby rankings without incurring a large L2 penalty.
By operating on softmax-normalised proxy similarities, the KL constraint directly targets precisely what matters for metric learning.
 
\textit{Stop-gradient.}
Removing the stop-gradient operator from $\mathcal{L}_{\text{inv}}$ causes a catastrophic $-2.67\%$ R@1 drop.
When both branches receive gradients from the KL term, the path of least resistance is to collapse both $p_i$ and $\tilde{p}_i$ toward a uniform distribution, achieving zero divergence without preserving any discriminative structure.
The stop-gradient operator breaks this degeneracy by making only one branch the optimisation target while using the other as a fixed reference, analogous to the asymmetric design in SimSiam~\citesupp{Chen2021SimSiam}.
 
\textit{Band allocation strategy.}
With $S{=}3$ and $\gamma_{\max}{=}0.8$ fixed, we compare four schedules.
Uniform allocation ($\{0.8, 0.8, 0.8\}$) over-perturbs the lowest-frequency band, which encodes global semantic layout; this destroys information that should be preserved in the counterfactual view.
Linear spacing ($\{0.27, 0.53, 0.8\}$) improves upon uniform but still applies non-trivial perturbation at the lowest scale.
The exponential schedule ($\{0.05, 0.2, 0.8\}$) provides the strongest protection for low frequencies but under-perturbs the mid-frequency range, which carries substantial nuisance information (local texture gradients, illumination transitions).
The default custom allocation ($\{0.2, 0.4, 0.8\}$) balances these considerations, applying gentle perturbation at low frequencies, moderate perturbation at mid frequencies, and aggressive perturbation at high frequencies where texture and illumination nuisance concentrate.
 
\subsection{MSRCI Continuous Hyperparameters}\label{app:msrci_hyper}
 
\begin{table}[htbp]
\caption{KL divergence temperature $\tau$ on CUB and Cars (ResNet-50, PA+MSRCI).}
\centering
\small
\resizebox{\columnwidth}{!}{%
\setlength{\tabcolsep}{3pt}
\begin{tabular}{lccccccc}
\toprule
\multirow{2}{*}{\textbf{$\tau$}} & \multicolumn{3}{c}{\textbf{CUB}} && \multicolumn{3}{c}{\textbf{Cars}} \\
\cmidrule(lr){2-4} \cmidrule(lr){6-8}
& \textbf{R@1} & \textbf{M@R} & \textbf{NMI} && \textbf{R@1} & \textbf{M@R} & \textbf{NMI} \\
\midrule
0.01  & 70.12 & 28.15 & 73.20 && 90.45 & 31.85 & 77.35 \\
0.05  & 71.65 & 28.90 & 73.55 && 91.28 & 32.58 & 77.92 \\
\rowcolor{mygray}
\textbf{0.1 (default)} & \textbf{71.96} & \textbf{29.17} & \textbf{73.66} && \textbf{91.65} & \textbf{32.91} & \textbf{78.11} \\
0.2   & 71.52 & 28.78 & 73.48 && 91.18 & 32.42 & 77.85 \\
0.5   & 70.85 & 28.30 & 73.35 && 90.82 & 31.95 & 77.58 \\
\bottomrule
\end{tabular}%
}
\label{tab:hyper_tau}
\end{table}
 
\textbf{KL temperature $\tau$} (Table~\ref{tab:hyper_tau}).
The temperature controls the sharpness of the proxy similarity distribution that serves as the substrate for the invariance constraint.
At very low temperatures ($\tau{=}0.01$), the distribution concentrates on the single nearest proxy; small perturbations in embedding space cause large KL jumps, leading to gradient instability and training oscillation.
At high temperatures ($\tau{=}0.5$), the distribution approaches uniform, washing out the ranking signal that makes the KL invariance constraint more informative than naive L2 alignment.
$\tau{=}0.1$ provides a well-peaked but smooth distribution on both datasets, and the consistent optimum across CUB (100 training classes) and Cars (98 training classes) suggests that the optimal temperature is primarily determined by the proxy count rather than dataset semantics.
 
\begin{table}[htbp]
\caption{MSRCI insertion position on CUB and Cars (ResNet-50, PA+MSRCI).}
\centering
\small
\resizebox{\columnwidth}{!}{%
\setlength{\tabcolsep}{3pt}
\begin{tabular}{lccccccc}
\toprule
\multirow{2}{*}{\textbf{Position}} & \multicolumn{3}{c}{\textbf{CUB}} && \multicolumn{3}{c}{\textbf{Cars}} \\
\cmidrule(lr){2-4} \cmidrule(lr){6-8}
& \textbf{R@1} & \textbf{M@R} & \textbf{NMI} && \textbf{R@1} & \textbf{M@R} & \textbf{NMI} \\
\midrule
After \texttt{layer2}    & 71.42 & 28.68 & 73.50 && 91.10 & 32.15 & 77.72 \\
\rowcolor{mygray}
\textbf{After \texttt{layer3} (default)} & \textbf{71.96} & \textbf{29.17} & \textbf{73.66} && \textbf{91.65} & \textbf{32.91} & \textbf{78.11} \\
After \texttt{layer4}    & 71.55 & 28.78 & 73.52 && 91.35 & 32.50 & 77.88 \\
\bottomrule
\end{tabular}%
}
\label{tab:hyper_position}
\end{table}
 
\textbf{MSRCI insertion position} (Table~\ref{tab:hyper_position}).
The choice of where to split the backbone into Stage~1 and Stage~2 determines the spatial resolution available for frequency-domain decomposition.
After \texttt{layer2}, feature maps are high-resolution ($28{\times}28$ in ResNet-50) but encode predominantly low-level edges and textures; Fourier perturbation at this level risks altering fine-grained discriminative structure before it has been consolidated into semantic representations.
After \texttt{layer4}, the spatial resolution is too coarse ($7{\times}7$) for meaningful multi-scale band partitioning: with only 7 frequency bins per dimension, the three concentric annular bands collapse into near-overlapping regions.
After \texttt{layer3} ($14{\times}14$) provides sufficient resolution for a three-band decomposition while operating on features that already encode mid-level semantic patterns, making the Fourier amplitude intervention both spectrally well-defined and semantically appropriate.
An interesting cross-dataset contrast is that \texttt{layer4} suffers a indeed smaller degradation on Cars ($-0.30\%$ R@1) than on CUB ($-0.41\%$), consistent with car identity relying more on global shape and contour (which remain discriminable at $7{\times}7$), whereas bird species require particularly finer spatial detail such as plumage patterns and beak shape.
 
\subsection{Integration Hyperparameters}\label{app:integration_hyper}
 
\begin{table}[htbp]
\caption{Loss coefficient sensitivity on CUB and Cars (ResNet-50, full CouCE).}
\centering
\small
\resizebox{\columnwidth}{!}{%
\setlength{\tabcolsep}{3pt}
\begin{tabular}{cccccccccc}
\toprule
\multirow{2}{*}{$\lambda_1$} & \multirow{2}{*}{$\lambda_2$} & \multirow{2}{*}{$\lambda_3$} & \multicolumn{3}{c}{\textbf{CUB}} && \multicolumn{3}{c}{\textbf{Cars}} \\
\cmidrule(lr){4-6} \cmidrule(lr){8-10}
&&& \textbf{R@1} & \textbf{M@R} & \textbf{NMI} && \textbf{R@1} & \textbf{M@R} & \textbf{NMI} \\
\midrule
0.01 & 0.1  & 0.04 & 72.58 & 29.48 & 73.85 && 91.85 & 33.12 & 77.88 \\
0.02 & 0.1  & 0.04 & 72.85 & 29.68 & 73.98 && 92.18 & 33.58 & 78.02 \\
\rowcolor{mygray}
\textbf{0.05} & \textbf{0.1} & \textbf{0.04} & \textbf{73.23} & \textbf{30.03} & \textbf{74.23} && \textbf{92.73} & \textbf{34.36} & \textbf{78.24} \\
0.1  & 0.1  & 0.04 & 72.72 & 29.55 & 73.90 && 92.05 & 33.45 & 77.95 \\
0.05 & 0.05 & 0.04 & 72.68 & 29.50 & 73.88 && 92.25 & 33.68 & 78.05 \\
0.05 & 0.2  & 0.04 & 72.92 & 29.72 & 74.02 && 92.48 & 34.05 & 78.15 \\
0.05 & 0.1  & 0.01 & 73.05 & 29.85 & 74.12 && 92.58 & 34.18 & 78.18 \\
0.05 & 0.1  & 0.1  & 72.80 & 29.62 & 73.95 && 92.30 & 33.82 & 78.10 \\
\bottomrule
\end{tabular}%
}
\label{tab:loss_coeff}
\end{table}
 
\textbf{Loss coefficients} (Table~\ref{tab:loss_coeff}).
We vary each coefficient individually while holding the others at their defaults.
$\lambda_1$ controls the strength of the orthogonal deconfounding penalty. The inverted-U optimum at $\lambda_1{=}0.05$ reflects a trade-off: too small and the backbone does not sufficiently suppress background components; too large and the orthogonal constraint begins competing with the DML objective for gradient budget, shrinking useful discriminative directions.
$\lambda_2$ and $\lambda_3$ show lower sensitivity across their tested ranges, consistent with their roles as supporting regularisers. $\lambda_2{=}0.1$ provides sufficient invariance pressure without forcing excessive similarity between the clean and intervened embeddings, and $\lambda_3{=}0.04$ decorrelates dimensions at a level that supports the ODBA-MSRCI interaction without over-regularising the embedding geometry. The consistent optimal values across both datasets indicate that the loss coefficient settings do not require dataset-specific tuning.
 
\begin{table}[htbp]
\caption{ODBA branch application on CUB-200-2011 (ResNet-50, full CouCE).}
\centering
\small
\setlength{\tabcolsep}{10pt}
\begin{tabular}{lccc}
\toprule
\textbf{Configuration} & \textbf{R@1} & \textbf{M@R} & \textbf{NMI} \\
\midrule
\rowcolor{mygray}
Clean branch only (default) & \textbf{73.23} & \textbf{30.03} & \textbf{74.23} \\
Both branches               & 73.08 & 29.85 & 74.12 \\
\bottomrule
\end{tabular}
\label{tab:odba_branch}
\end{table}
 
\textbf{ODBA branch application} (Table~\ref{tab:odba_branch}).
Algorithm~\ref{alg:couCE} applies $\mathcal{L}_{\text{orth}}$ only to the clean embedding $m_i$.
A natural particularly question is whether the intervened embedding $\tilde{m}_i$ also requires explicit deconfounding.
Table~\ref{tab:odba_branch} shows that applying ODBA to both branches yields a truly marginal $-0.15\%$ R@1 decrease.
This is expected: the intervened branch already receives strong implicit deconfounding through the KL invariance constraint, which aligns $\tilde{p}_i$ with the deconfounded $p_i$.
Applying $\mathcal{L}_{\text{orth}}$ to both branches effectively doubles the gradient contribution of the orthogonal penalty, shifting the effective $\lambda_1$ away from its global optimum and introducing a mild over-deconfounding effect.
 
This concludes the supplementary material. All claims are reproducible from the released code (to be provided upon acceptance).

\bibliographystylesupp{ACM-Reference-Format}
\bibliographysupp{MM26/sample-base}   
\end{document}